\documentclass[journal, twoside]{IEEEtran}

\pdfoutput=1

\usepackage{amssymb}
\usepackage{booktabs}     
\usepackage{multirow}
\usepackage{ulem}         
\usepackage{graphicx}
\usepackage{pifont}
\usepackage{array}
\usepackage{subfigure}    
\usepackage{hhline}
\usepackage{color}
\usepackage{epstopdf}
\usepackage{diagbox}
\usepackage[colorlinks,linkcolor=black]{hyperref}

\graphicspath{{figures/}}

\ifCLASSINFOpdf
\else
\fi
\usepackage{amsmath}
\usepackage{array}
\usepackage{url}

\newcommand{\PreserveBackslash}[1]{\let\temp=\\#1\let\\=\temp}
\newcolumntype{C}[1]{>{\PreserveBackslash\centering}p{#1}}
\newcolumntype{R}[1]{>{\PreserveBackslash\raggedleft}p{#1}}
\newcolumntype{L}[1]{>{\PreserveBackslash\raggedright}p{#1}}

\begin{document}
%
\title{Channel Splitting Network for Single MR Image Super-Resolution}
%

\author{Xiaole~Zhao,~Yulun~Zhang,~Tao~Zhang,~and~Xueming~Zou
\thanks{Manuscript received  October 12, 2018; revised March 17, 2019; accepted May 30, 2019. Date of publication XX XX, 2019; date of current version XX XX, 2019. This work is supported in part by Sichuan Science and Technology Program under grant 2019YJ0181, and in part by National Key Research and Development Program of China under grants 2016YFC0100800 and 2016YFC0100802. \textit{(Corresponding author: Tao Zhang.)}}
\thanks{X. Zhao is with the School of Life Science and Technology, University of Electronic Science and Technology of China (UESTC), Chengdu, Sichuan 611731, China (e-mail: zxlation@foxmail.com).}
\thanks{T. Zhang and X. Zou are with the High Field Magnetic Resonance Brain Imaging Laboratory of Sichuan and Key Laboratory for NeuroInformation of Ministry of Education, Chengdu, Sichuan 611731, China; They are also with the School of Life Science and Technology, University of Electronic Science and Technology of China (UESTC), Chengdu, Sichuan 611731, China (e-mail: taozhangjin@gmail.com; mark.zou@alltechmed.com).}
\thanks{Y. Zhang is with the Department of Electrical and Computer Engineering, Northeastern University, Boston, MA 02115, USA (e-mail: yulun100@gmail.com).}
\thanks{Digital Object Identifier XXXX}}

%
%

\markboth{}
{Zhao \MakeLowercase{\textit{et al.}}: Channel Splitting Network for Single MR Image Super-Resolution}
%



\maketitle

\begin{abstract}
High resolution magnetic resonance (MR) imaging is desirable in many clinical applications due to its contribution to more accurate subsequent analyses and early clinical diagnoses. Single image super-resolution (SISR) is an effective and cost efficient alternative technique to improve the spatial resolution of MR images. In the past few years, SISR methods based on deep learning techniques, especially convolutional neural networks (CNNs), have achieved state-of-the-art performance on natural images. However, the information is gradually weakened and training becomes increasingly difficult as the network deepens. The problem is more serious for medical images because lacking \textit{high quality} and \textit{effective} training samples makes deep models prone to underfitting or overfitting. Nevertheless, many current models treat the hierarchical features on different channels equivalently, which is not helpful for the models to deal with the hierarchical features discriminatively and targetedly. To this end, we present a novel channel splitting network (CSN) to ease the representational burden of deep models. The proposed CSN model divides the hierarchical features into two branches, i.e., residual branch and dense branch, with different information transmissions. The residual branch is able to promote feature reuse, while the dense branch is beneficial to the exploration of new features. Besides, we also adopt the merge-and-run mapping to facilitate information integration between different branches. Extensive experiments on various MR images, including proton density (PD), T1 and T2 images, show that the proposed CSN model achieves superior performance over other state-of-the-art SISR methods.

\end{abstract}


%
\IEEEpeerreviewmaketitle

\section{Introduction}
%
%
%
%
\IEEEPARstart{S}{patial} resolution is one of the most important imaging parameters for magnetic resonance imaging (MRI). In many clinical applications and research work, high resolution (HR) MRI is usually preferred because it can provide more significant structure and texture details with a smaller voxel size \cite{Carmia2006Resolution}, thus promoting accurate subsequent analysis and early diagnosis. However, it is limited by several factors, e.g., hardware device, imaging time, desired signal-to-noise ratio (SNR) and body motion etc, and increasing spatial resolution of magnetic resonance (MR) images typically reduces image SNR and/or increases imaging time \cite{Plenge2012Super}.

\begin{figure}[t]
  \centering
  \subfigure[Validation curves]{\label{fig1a}
  \begin{minipage}[t]{0.23\textwidth}
    \centering
    \includegraphics[scale = 0.28]{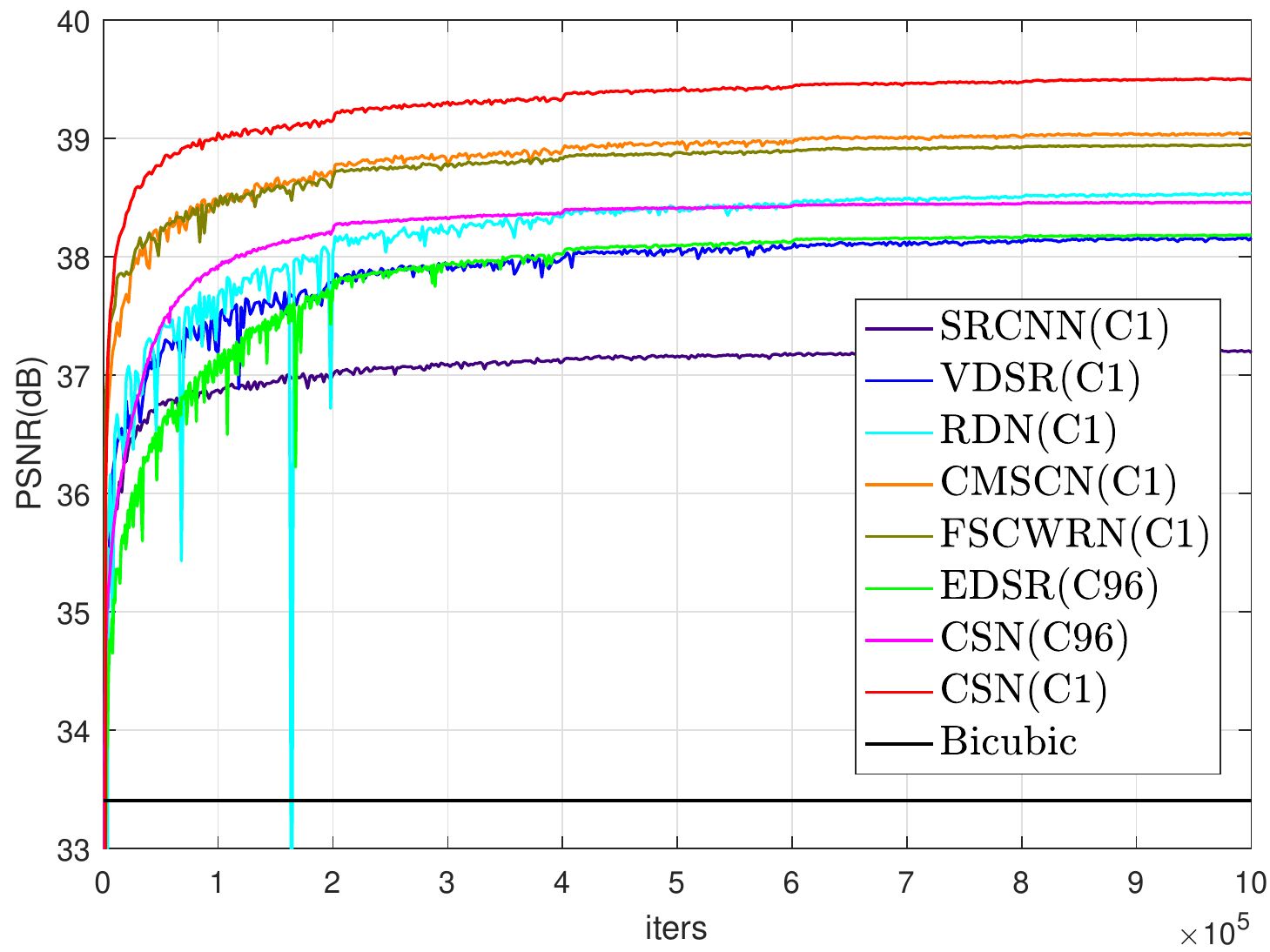}
  \end{minipage}}
  \subfigure[Testing results]{\label{fig1b}
  \begin{minipage}[t]{0.23\textwidth}
    \centering
    \includegraphics[scale = 0.28]{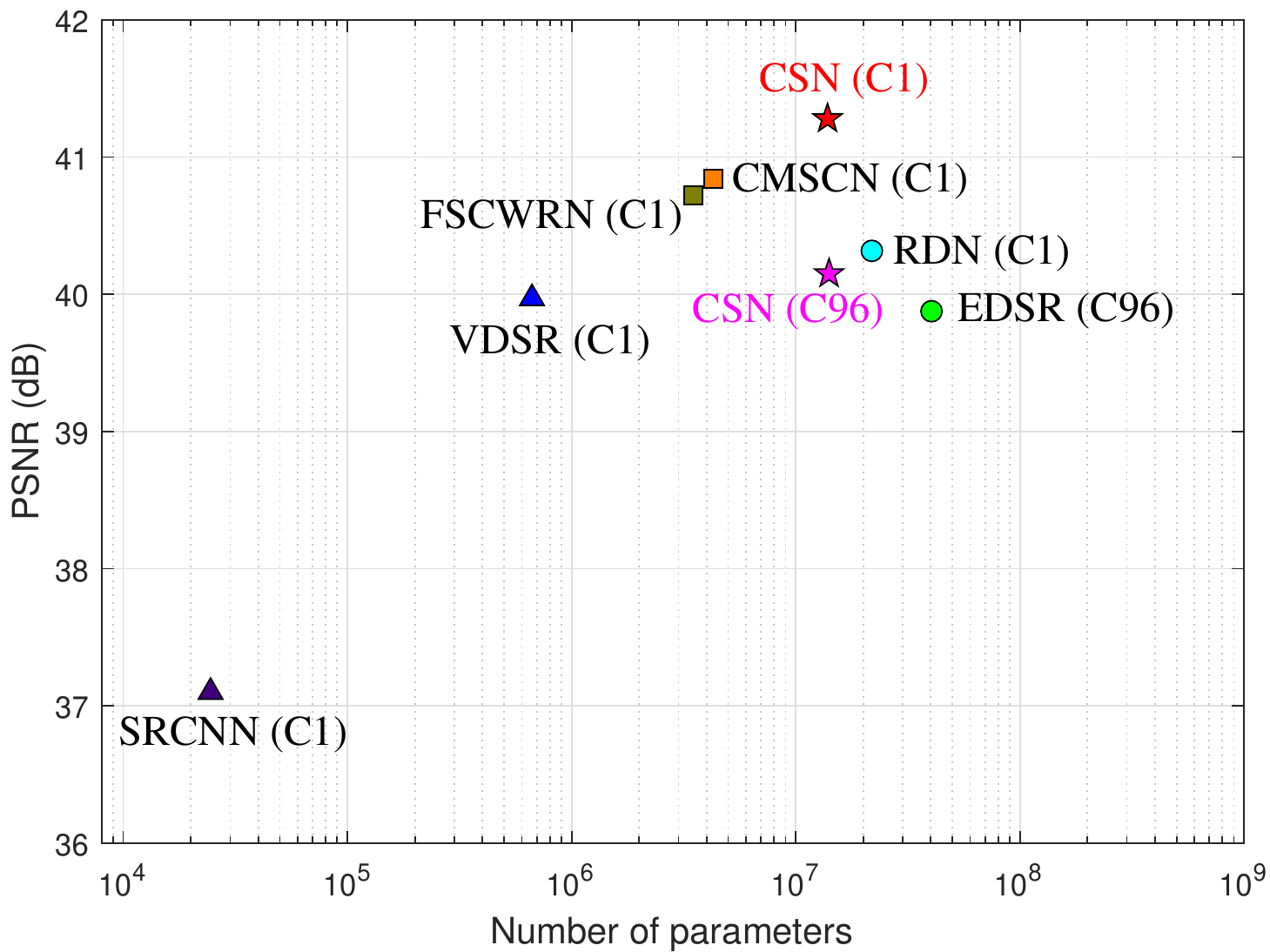}
  \end{minipage}}
  \vspace{-2mm}
  \caption{The performance comparison between several SISR models on proton density (PD) volumes of IXI dataset for SR$\times$2. (a) The validation results on 6 PD volumes (576 2D slices). (b) The test results vs. the number of model parameters on 70 PD volumes (6720 2D slices, Bicubic: 35.04 dB). The symbols $\vartriangle$, {\tiny $\square$}, $\text{\ding{73}}$ and $\circ$ represent models with less than 1M, 10M, 20M and more than 20M parameters respectively. C1 indicates that the training sample is a single slice, and C96 indicates that the model treats 96 slices of a 3D volume as 96 channels.}\label{fig1}
\end{figure}

Image super-resolution (SR) is a typical ill-posed inverse problem in computer vision community, which mainly aims at inferring a HR image from one or more low resolution (LR) images. It is a well-studied problem in both natural image (NI) and MR image processing. High resolution means that the pixel density of an image is higher than its LR counterpart. Thus, HR images can offer more details that may be critical in various applications such as medical imaging \cite{Hu2016Single, Shi2013Cardiac}, aerial spectral imaging \cite{Rangnekar2017Aerial} and remote sensing imaging \cite{Thornton2006Subpixel, Pan2013Super} and security and surveillance \cite{Ahmad2012An}, where high frequency details are very important and greatly desired. Up to now, many SR methods have been studied and proposed. Early methods include: (i) interpolation methods, e.g., bicubic, Lanczos-$\sigma$ \cite{Gottlieb1997On}; (ii) modeling and reconstruction methods, e.g., iterative back projection (IBP) \cite{Irani1991Improving}, projection onto convex set (POCS) \cite{Stark1989High} etc.; (iii) traditional shallow learning methods, e.g., example learning \cite{Freeman2002Example,Kim2013Example}, dictionary learning \cite{Yang2008Image, Yang2012Coupled} etc. The performance of these methods is inherently limited because the additional information available for solving this ill-posed inverse problem is also very limited, e.g., the interpolation methods make use of the basic smoothing priori by implicitly assuming that the image signal is continuous and bandwidth limited, and traditional machine learning-based methods can learn insufficient information due to the limited representational capacity of these shallow models.

\begin{figure*}[t]
  \centering
  \includegraphics[width = \textwidth]{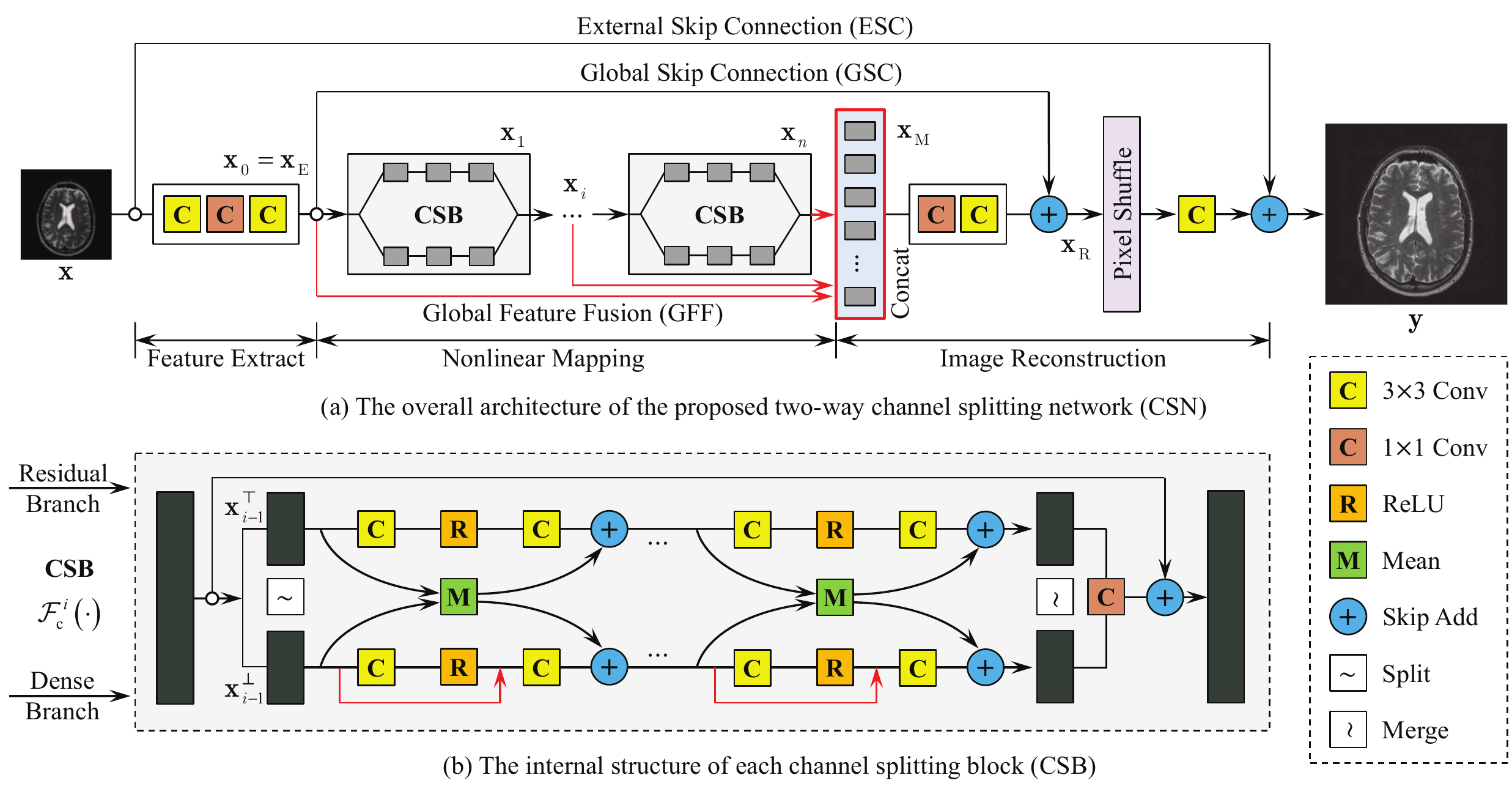}   \\
  \vspace{-2mm}
  \caption{The diagram of the proposed CSN model. (a) The overall structure consists of three parts: feature extraction $\mathcal{F}_{\text{E}}(\cdot)$, nonlinear mapping $\mathcal{F}_{\text{M}}(\cdot)$ and image reconstruction $\mathcal{F}_{\text{F}}(\cdot) + \mathcal{F}_{\text{R}}(\cdot)$. (b) Channel splitting block (CSB). The intermediate feature maps within a CSB are split into two branches along the channel direction. One is built as a residual-like structure (residual branch, top), and the other is built as a dense-like structure (dense branch, bottom). {\textcolor[rgb]{1,0,0}{Red}} arrows in GFF \cite{Zhang2018Residual} and dense branch indicate dense connection (channel concatenation).}\label{fig:overall_structure_CSN}
\end{figure*}

In recent years, various advanced SR methods have emerged with the rapid development of deep learning techniques \cite{Lecun2015Deep} and greatly promoted the best state of SR performance. Super-resolution convolutional neural network (SRCNN) \cite{Dong2016Image} and fast super-resolution convolutional neural network (FSRCNN) \cite{Dong2016Accelerating} are two pioneering contributions that utilize convolutional neural networks (CNNs) \cite{Lecun1998Gradient} to solve SR tasks. The further improvement based on these pioneering work mainly focused on increasing model depth or sharing model parameters at the beginning, for example, deeply recursive convolutional network (DRCN) \cite{Kim2016Deeply}, deep recursive residual network (DRRN) \cite{Tai2017Image}, super-resolution using very deep convolutional networks (VDSR) \cite{Kim2016Accurate} and memory network (MemNet) \cite{Tai2017Memnet} etc. These methods, however, are mainly aimed at the SR task of natural images, not specially at medical images.

The medical image processing community has noticed these advances and some medical image SR methods based on deep learning have also appeared \cite{Pham2017Brain,Chen2018Brain,Can2018Self,Chen2018Efficient,Shi2019MR,Shi2018Super}. For MR images, training deep models with a large amount of parameters and extremely deep structures is usually more difficult because \textit{high-quality} and \textit{effective} training samples are relatively scarce and unavailable. It is worth noting that \textit{the challenge is not the availability of training data itself, but the acquisition or the quality of relevant annotations/labeling for these data} \cite{Litjens2017A}. Therefore, some SR models that aim at NI are likely to fail when directly trained with MR images despite sufficient training samples, e.g., Fig.\ref{fig1} displays the peak signal-to-noise ratio (PSNR) performance of several recent single image super-resolution (SISR) models and the proposed channel splitting network (CSN) on proton density (PD) images of IXI dataset (\url{http://brain-development.org/ixi-dataset/}) for SR$\times$2, where enhanced deep super-resolution network (EDSR) \cite{Lim2017Enhanced} and residual dense network (RDN) \cite{Zhang2018Residual} are advanced models on NI. But it is failed to train the EDSR model (the same configuration as \cite{Lim2017Enhanced}) with 48000 2D PD images. This problem of training failure caused by the degradation of sample quality will get worse as the network depth (or width) and the number of model parameters increase.

Thus, in the context of MR image super-resolution based on deep learning techniques, the dilemma has become more apparent: \textit{on the one hand, models with shallow structures and fewer parameters are easy to train, but their SR performance is usually unsatisfactory; on the other hand, models with deeper structures and more parameters are promising to improve SR performance, but it is more difficult for them to be fully trained with MR images}. An effective way to alleviate the difficulty of model training is residual learning, which is initially proposed for image recognition \cite{He2016Deep,He2015Identity}. It has been widely proved to be helpful for feature reuse and model convergence, thus making it possible to build extremely deep models. However, residual learning strategy alone is still insufficient to train the model with a very deep structure and a very large number of parameters in case of MR images, e.g., EDSR \cite{Lim2017Enhanced} with about 43M parameters is a typical residual network, but the original configuration can hardly be well trained with 2D MR images in our settings. The problem of training failure can be addressed by concatenating multiple 2D MR images into a single multi-channel training sample at the expense of performance, e.g., we train the original EDSR \cite{Lim2017Enhanced} model by taking all 96 slices of a 3D volume as 96 channels of a single training sample, as shown in Fig.\ref{fig1} (marked as EDSR (C96)).

In this paper, we improve the above dilemma by introducing a \textit{deep channel splitting network (CSN)} framework. It assumes that the hierarchical features of deep models have certain clustering properties, and explicitly discriminating them is beneficial to ease the representational burden of deep models and further improve the SR performance. Therefore, instead of transferring the feature maps of the previous layer completely to the next layer, we split the feature maps into two different parts (branches) with different information transmissions. Each branch can be structured differently, e.g., in this work, we use propagation mechanisms similar to residual network (ResNet) \cite{He2016Deep,He2015Identity} and dense network (DenseNet) \cite{Huang2016Densely} (or RDN \cite{Zhang2018Residual}) on each branch. Besides, the merge-and-run (MAR) mapping \cite{Zhao2017Deep,Hu2018Single} is also applied to facilitate the information integration of different branches. Thus, our model has two notable characteristics: \textit{(1) channel splitting discriminatorily limits the hierarchical features into different clusters and reduces the representational redundancy of the model by curtailing the internal connections; (2) the merge-and-run mapping can promote information sharing and integration between the hierarchical features and therefore help to improve the information flow through the entire network}.

To make full use of the hierarchical features, we also adopt the global feature fusion (GFF) technique proposed by \cite{Zhang2018Residual}, as shown in Fig.\ref{fig:overall_structure_CSN}(a). Moreover, multilevel residual mechanism and constant scaling technique \cite{Lim2017Enhanced,Szegedy2016Inception} are also applied to our models to further stabilize the model training. To verify the effectiveness of the proposed model, a set of standard datasets for the task of single MR image SR is generated from the IXI dataset, which includes three types of MR images (i.e., PD, T1 and T2, each of which can exhibit different contrasts for the same image content.) and two kinds of degeneration (bicubic downsampling and $k$-space truncation). The quantitative and qualitative experiments on the datasets display the superiority of the proposed model over other advanced methods.

The rest of this paper is organized as follows. In section \ref{sec:relatedwork}, we present some previous contributions related the present work. The proposed method and the experimental results are detailed in section \ref{sec:proposedmethod} and section \ref{sec:experiments}, respectively. Section \ref{sec:discussion} gives some discussion and future work. Finally, we conclude the whole work in section \ref{sec:conclusion}.

\section{Related Work}
\label{sec:relatedwork}
\subsection{Super-Resolution with Deep Learning}
\begin{figure}[t]
  \centering
  \includegraphics[width = 0.48\textwidth]{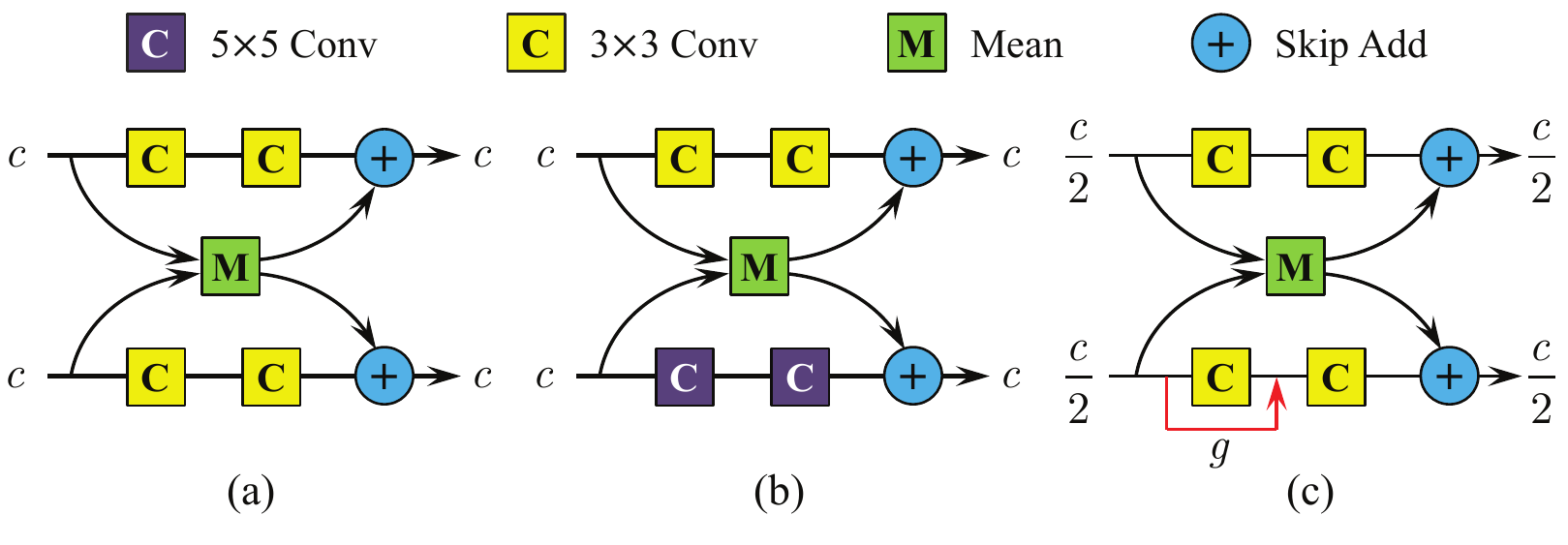}   \\
  \vspace{-2mm}
  \caption{The merge-and-run stage mappings in a building block. (a) Original mapping \cite{Zhao2017Deep}. (b) Multi-scale cross (MSC) mapping \cite{Hu2018Single}. (c) The proposed mapping. $c$ is the channel number of feature maps, and $g$ is the growth of a dense connection \cite{Huang2016Densely}. BN and ReLU are omitted for simplification, and the red arrow indicates the dense connection.}\label{fig3}
\end{figure}

Although the work using artificial neural networks (ANNs) to solve SR problems has emerged as early as 2006 \cite{Huang2006Super}, the pioneering work with deep learning techniques in the modern sense is SRCNN \cite{Dong2016Image}. Subsequently, some advanced methods based on the increase of network depth and parameter sharing are proposed. Kim \textit{et al.} \cite{Kim2016Accurate} increased network depth by stacking multiple conv layers and the usage of global residual learning (GRL), and firstly introduced recursive learning trick in a deep network for parameter sharing \cite{Kim2016Deeply}. Another network introduced by Tai \textit{et al.} \cite{Tai2017Image} has utilized recursive blocks to reuse parameters. Motivated by the fact that human thoughts have persistency, a deep persistent memory network (MemNet) which consists of the so-called memory block, has also been proposed by the same author \cite{Tai2017Memnet}. To improve information flow and capture more sufficient knowledge for reconstructing the high frequency details, Hu \textit{et al.} \cite{Hu2018Single} proposed a cascaded multi-scale cross network (CMSCN) in which a sequence of subnetworks (Fig.\ref{fig3}(b)) is cascaded to infer HR features in a coarse-to-fine manner. To some extent, these methods promote the design of new structures for image generation tasks.

There is a common feature among the above methods: they use the bicubic-interpolated image of the original LR image as input to their models. This preprocessing is convenient for keeping the size of the output image consistent with the target HR image. However, it places the nonlinear inference of the network in HR image space, resulting in great computation and memory consumption. There exist two solutions for this issue currently: deconvolution or transpose convolution \cite{Dong2016Accelerating}, and the efficient sub-pixel convolutional neural network (ESPCNN) \cite{Shi2016Real}. Both of them can effectively solve the above problem by shifting the HR reference to the LR image space. Benefiting from the nonlinear mapping within the LR image space, these methods are capable of increasing the scale of deep models and thus boost the SR performance greatly, e.g., EDSR/MDSR \cite{Lim2017Enhanced}. Further, Zhang \textit{et al.} \cite{Zhang2018Residual} combined the idea of residual learning \cite{He2016Deep,He2015Identity} with densely connected DenseNet \cite{Huang2016Densely} and proposed a novel residual dense network (RDN) to fully utilize the hierarchical features of deep models.

Contrary to the pursuit of high performance, some methods aim to improve the tradeoff between SR performance and time efficiency to improve the practicality of the model, e.g., \cite{Shi2016Real,Yamanaka2017Fast,Ahn2018Fast}. Overall, although these methods favor relatively fast inference, they are still lightweight and small-scale so that their representational capacity is limited to some extent.

\begin{figure*}[t]
  \centering
  \subfigure[HR image]{\label{fig4a}
  \begin{minipage}[b]{0.234\textwidth}
    \centering
    \includegraphics[scale = 0.6]{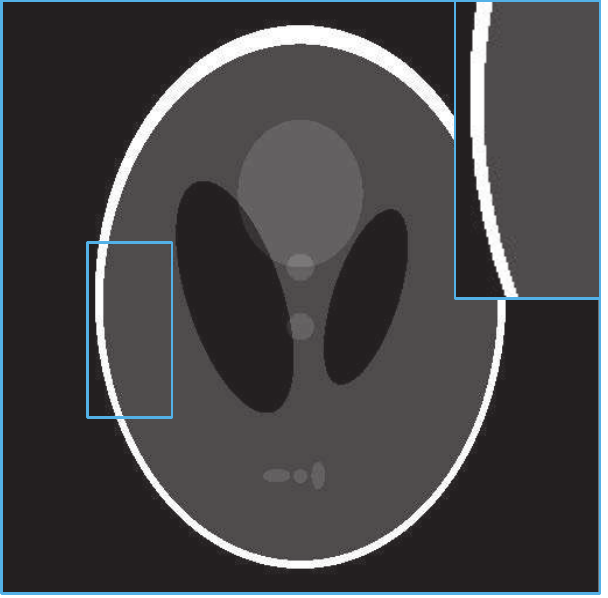}
  \end{minipage}}
  \subfigure[Fourier transform]{\label{fig4b}
  \begin{minipage}[b]{0.234\textwidth}
    \centering
    \includegraphics[scale = 0.6]{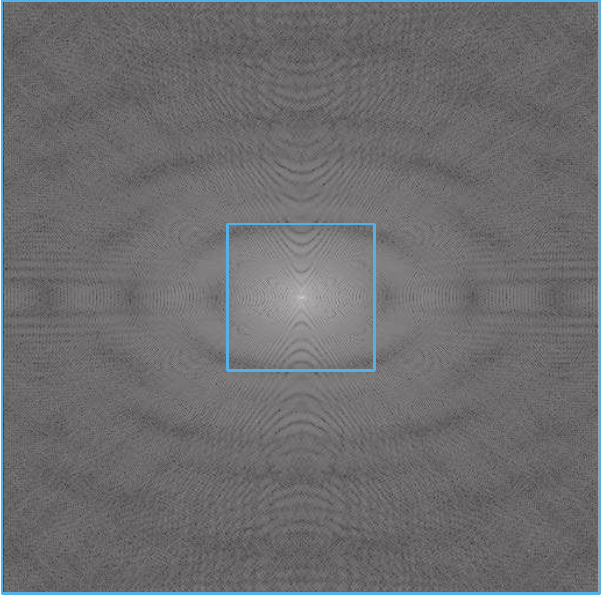}
  \end{minipage}}
  \hspace{-1.5mm}
  \subfigure[$k$-space truncation]{\label{fig4c}
  \begin{minipage}[b]{0.234\textwidth}
    \centering
    \includegraphics[scale = 0.6]{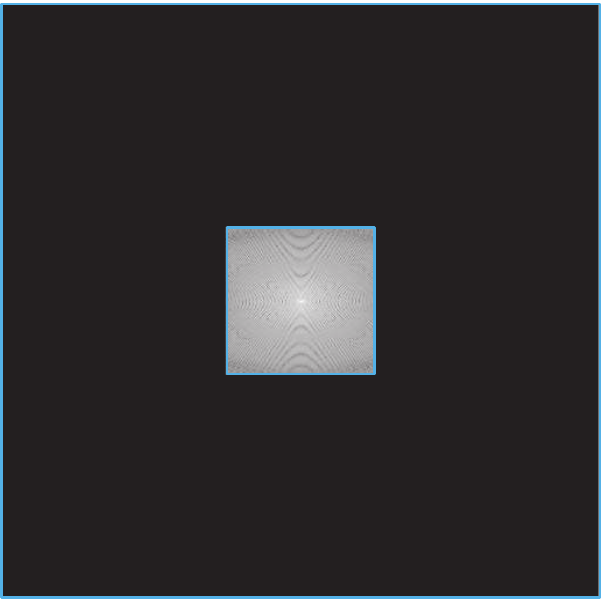}
  \end{minipage}}
  \subfigure[Inverse recovery]{\label{fig4d}
  \begin{minipage}[b]{0.234\textwidth}
    \centering
    \includegraphics[scale = 0.6]{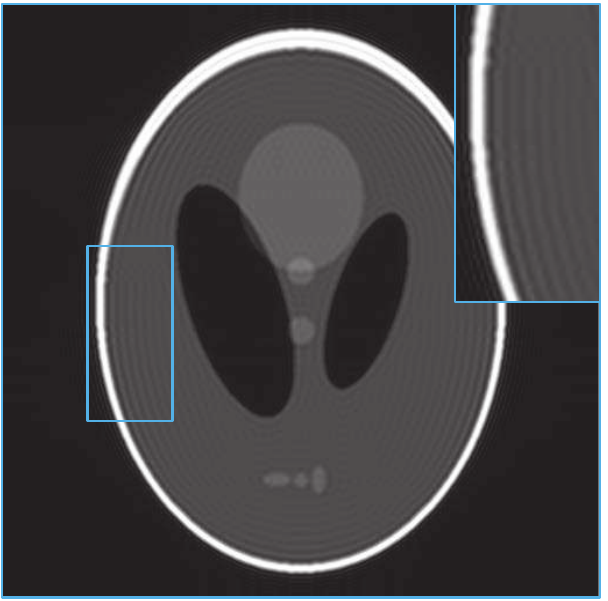}
  \end{minipage}}
  \caption{The illustration of $k$-space truncation degradation (TD) for SR$\times$4. Different from bicubic downsampling, the generated LR images are sometimes accompanied by Gibbs-ringing artifacts. The result in (d) is generated by zero-padding in $k$-space and inverse Fourier transform (IFT) for display purpose.}
  \label{fig4}
\end{figure*}

\subsection{MR Image Super-Resolution}
The early application of SR techniques to medical images mainly focuses on multi-frame image super-resolution (MFSR) tasks. For example, IBP \cite{Irani1991Improving} was adopted to generate a new image with increased spatial resolution from several spatially shifted, single-shot and diffusion-weighted brain MR images \cite{Peled2001Super}. Greenspan \textit{et al.} \cite{Greenspan2002MRI} and Shilling \textit{et al.} \cite{Shilling2009A} employed IBP and POCS \cite{Stark1989High} to produce a 3D MR volume with isotropic resolution from several 2D slices, respectively. These methods are usually accompanied with specific data acquisitions (e.g., rotation, scaling and translation) to simulate the generation of LR images. However, recovering a HR image from multiple degraded LR images usually needs to calibrate and fuse these LR images, which is a very challenging task in itself.

To avoid the difficulty of calibration and fusion between multiple LR images, Rousseau \cite{Rousseau2008Brain} first proposed to enhance MR image resolution with single image SR techniques. In this method, the extra information was introduced into the reconstruction process by referring to another HR image. A similar method was proposed by Manj$\acute{\rm o}$n \textit{et al.} \cite{Manj車n2010MRI}. They also used a HR image as reference but with a different strategy to produce HR images. These methods introduce very limited extra information because they learn knowledge from only one external HR image. SR methods based on conventional machine learning, e.g., sparse representation \cite{Rueda2013Single,Wang2014Sparse} and compressive sensing \cite{Roohi2012Super}, are also applied to medical images subsequently. Recently, more advanced SR methods based on deep learning \cite{Lecun2015Deep} have also been applied to MR image SR tasks \cite{Pham2017Brain,Chen2018Brain,Can2018Self,Chen2018Efficient,Oktay2016Multi}. However, these methods simply use deep learning techniques to deal with the SR tasks of MR images without considering the differences between natural images and medical images. Contrarily, the proposed CSN model aims at dealing with the hierarchical features discriminatively and reducing the representational burden of the model to adapt the degradation of MR training samples.

\subsection{Multi-Stream Networks}
Multi-stream networks are widely adopted in the image SR community to boost the model performance by assembling the information from different streams (paths). Wang \textit{et al.} \cite{Wang2016End} explored an end-to-end CNN architecture by jointly training both deep and shallow CNN networks, where the shallow one stabilizes model training and the deep one ensures an accurate HR reconstruction. Ren \textit{et al.} \cite{Ren2017Image} proposed a context-wise network fusion approach to integrate the outputs of individual networks by extra convolutional layers. Yamanaka \textit{et al.} \cite{Yamanaka2017Fast} combined skip connection layers and parallelized CNNs into a single CNN architecture. CMSCN \cite{Hu2018Single} is another multi-stream structure, in which complementary information under different receptive fields is integrated by the merge-and-run mechanism \cite{Zhao2017Deep} (Fig.\ref{fig3}). There are also multi-stream structures for medical image SR tasks, e.g., Oktay \textit{et al.} \cite{Oktay2016Multi} developed a multi-input cardiac image SR network, which is capable of assembling information from different viewing planes to improve the SR performance. These methods are fundamentally different from the proposed CSN model, in that they form the multi-stream structure by the reuse of the preceding features, while our CSN network construct the multi-stream structure by splitting the preceding features into different branches.

\begin{figure*}[t]
  \centering
  \includegraphics[width = 0.95\textwidth]{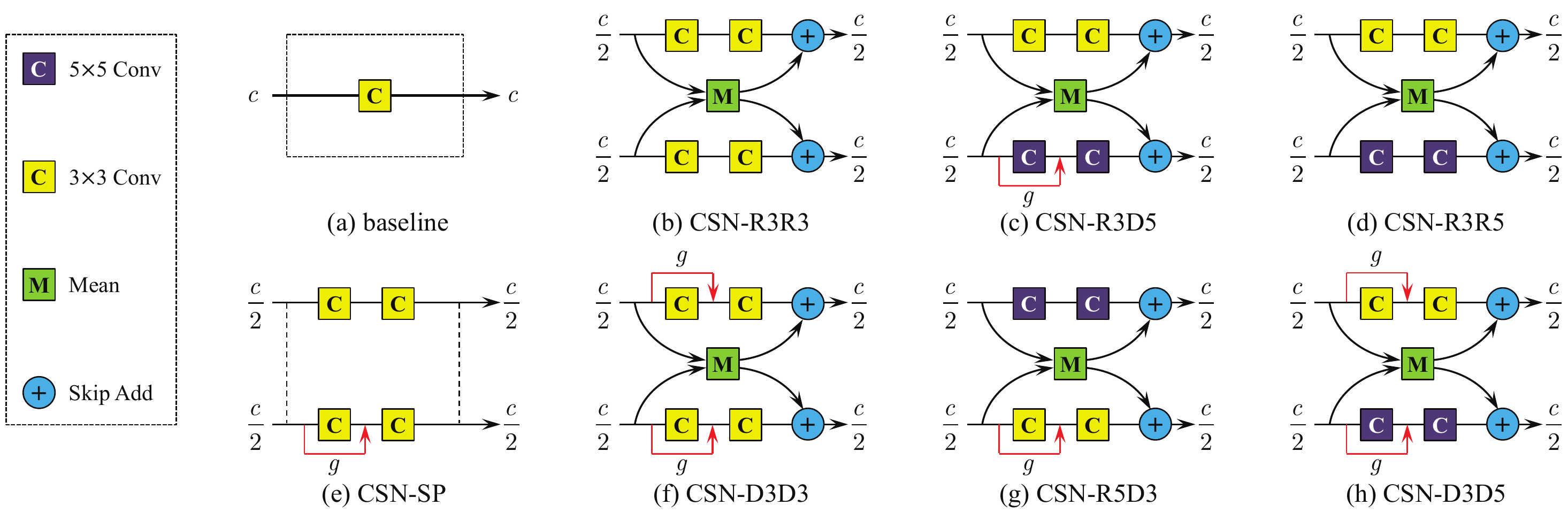}   \\
  \vspace{-2mm}
  \caption{Several stage mapping structures for comparison. The nonlinear function ReLU between two conv layers is omitted for simplification, and red arrows represent dense connection. Note Fig.\ref{fig3}(a) is different from CSN-R3R3 and there is a ReLU layer after the conv layer in (a). Please refer to Fig.\ref{fig:overall_structure_CSN}(b) for the overall structure of a CSB and the detailed structure of CSN-R3D3 stage mapping.}\label{fig5}
\end{figure*}

\section{Proposed Method}
\label{sec:proposedmethod}
\subsection{Overall Network Architecture}
The overall structure of the proposed CSN model is outlined in Fig.\ref{fig:overall_structure_CSN}. Similar to other deep models for image SR, it consists mainly of 3 parts: feature extraction, nonlinear mapping and image recovery. Firstly, the feature extraction network (FEN) is employed to express the input image $\mathbf{x}$ as a set of shallow features. These shallow features are then transmitted to the nonlinear mapping network (NMN), which contains a series of stacked channel splitting blocks (CSB). Subsequently, the hierarchical features from all CSBs are concatenated together to produce the final output of the NMN. This operation is also called global feature fusion (GFF) \cite{Zhang2018Residual}. Finally, the collected deep hierarchical features are fed into the image reconstruction network (IRN) to generate the final HR predication $\mathbf{y}$ of the entire CSN model.

\subsubsection{Feature Extraction}
The FEN contains two 3$\times$3 conv layers with a 1$\times$1 conv layer in the middle. Denote $\mathcal{F}_{\text{E}}(\cdot)$ as the corresponding mapping function, then the shallow features $\mathbf{x}_{\text{E}}$ extracted by the FEN can be represented as:
\begin{equation}
  \label{eqn1}
  \mathbf{x}_{\text{E}} = \mathcal{F}_{\text{E}}(\mathbf{x}),
\end{equation}
where $\mathbf{x}$ is the original LR input. The 1$\times$1 conv layer indicates a point-to-point linear transformation of the features extracted by the first 3$\times$3 conv layer. This 1$\times$1 conv layer is considered to be helpful to further improve the robustness of the extracted features because the features on different channels also contain spatial information in the context of image SR.

\subsubsection{Nonlinear Mapping}
The entire NMN net is denoted as $\mathcal{F}_{\text{M}}(\cdot)$. Therefore, the output of the NMN is given by $\mathcal{F}_{\text{M}}(\mathbf{x}_{\text{E}})$. Supposing we have $n$ CSBs in the entire network and $\mathbf{x}_{0} = \mathbf{x}_{E}$ is the input of the first CSB, then the output $\mathbf{x}_{i}$ of the $i$-th CSB can be obtained by:
\begin{equation}
  \label{eqn2}
  \mathbf{x}_{i} = \mathcal{F}_{\text{c}}^{i}(\mathbf{x}_{i-1}), \ \ i = 1,\ 2,\ \ldots,\ n,
\end{equation}
where the function $\mathcal{F}_{\text{c}}^{i}(\cdot)$ corresponds to the operations of the $i$-th CSB. More details about $\mathcal{F}_{\text{c}}^{i}(\cdot)$ will be presented in section \ref{subsec:csb}. Therefore, the output of the last CSB can be iteratively formulated as follow:
\begin{equation}
  \label{eqn3}
  \mathbf{x}_{n} = \mathcal{F}_{\text{c}}^{n}(\mathbf{x}_{n-1}) = \mathcal{F}_{\text{c}}^{n}(\mathcal{F}_{\text{c}}^{n-1}(\cdots(\mathcal{F}_{\text{c}}^{1}(\mathbf{x}_{0}))\cdots)).
\end{equation}
The output tensor of the $i$-th CSB $\mathbf{x}_{i}$ is produced by a series of operations (e.g., convolution, ReLU and constant scaling etc.) within the block, so it is viewed as a set of local feature maps \cite{Zhang2018Residual}. These local feature maps constitute the final output of our nonlinear mapping network. It should be noted that the output of the preceding CSB is directly used as the input of the next CSB. This is similar to the so-called continuous memory (CM) mechanism \cite{Zhang2018Residual} and contributes to the information propagation in the network \cite{He2015Identity}.

\subsubsection{Image Reconstruction}
This phase includes two related parts: global fusion of the local features $\mathbf{x}_{i}$ and the restoration of HR images based on the fused features. To fuse the local features, the outputs of all CSBs are first concatenated into a single tensor (red rectangle in Fig.\ref{fig:overall_structure_CSN}(a)):
\begin{equation}
  \label{eqn4}
  \mathbf{x}_{\text{M}} = [\mathbf{x}_{0}, \mathbf{x}_{1}, \ldots, \mathbf{x}_{n}],
\end{equation}
where $[\ldots]$ implies the concatenation. Next, the global features are extracted by fusing all local features from all the preceding channel splitting blocks. This is completed by a 1$\times$1 conv followed by a 3$\times$3 conv ($\mathcal{F}_{\text{F}}(\cdot)$ in Fig.\ref{fig:overall_structure_CSN}). Finally, the global residual learning (GRL) is used to stabilize the training of the model \cite{Lim2017Enhanced,Zhang2018Residual}, which is simply implemented via a global skip connection (GSC):
\begin{equation}
  \label{eqn5}
  \mathbf{x}_{\text{R}} = \mathcal{F}_{\text{F}}(\mathbf{x}_{\text{M}}) + \mathbf{x}_{0} = \mathcal{F}_{\text{F}}([\mathbf{x}_{0},\mathbf{x}_{1},\ldots,\mathbf{x}_{n}]) + \mathbf{x}_{0},
\end{equation}
where $\mathbf{x}_{\text{R}}$ is the fused features that will be used to recover the HR image $\mathbf{y}$. As for HR image restoration, it is mainly made up of a pixel shuffle layer followed by a 3$\times$3 convolutional layer, and an external residual learning (ERL). Formally, it can be represented as:
\begin{equation}
  \label{eqn6}
  \mathbf{y} = \mathcal{F}_{\text{R}}(\mathbf{x}_{\text{R}}) + \mathbf{x},
\end{equation}
where $\mathcal{F}_{\text{R}}(\cdot)$ is the function corresponding to the pixel shuffle layer and the following 3$\times$3 convolutional layer, and $\mathbf{x}$ is the original LR input to the model. Note that the pixel shuffle layer is implemented by ESPCNN \cite{Shi2016Real} in the way of \cite{Lim2017Enhanced}.

\subsection{Channel Splitting Block}
\label{subsec:csb}
The CMSCN network explored by \cite{Hu2018Single} is a multi-stream structure that integrates the complementary information under different receptive fields. Moreover, it has been proved that residual learning enables feature reuse and dense learning enables new features exploration, both of which are important for learning good representations \cite{Chen2017Dual}. Inspired by this, we present a two-way channel splitting block (CSB) to incorporate different information with different propagation mechanisms. As shown in Fig.\ref{fig:overall_structure_CSN}(b), the distinctive features of the proposed CSB are \textit{channel splitting and merging}, and \textit{fusion of residual learning and dense learning}. Besides, local residual learning (LRL) is also applied to further improve the information propagation. It has been shown that LRL is helpful to stabilize the training process and improve the representational capacity of the model, resulting better SR performance \cite{Zhang2018Residual}.

\begin{figure*}[t]
  \centering
  \subfigure[]{\label{fig6a}
  \begin{minipage}[t]{0.32\textwidth}
    \centering
    \includegraphics[scale = 0.38]{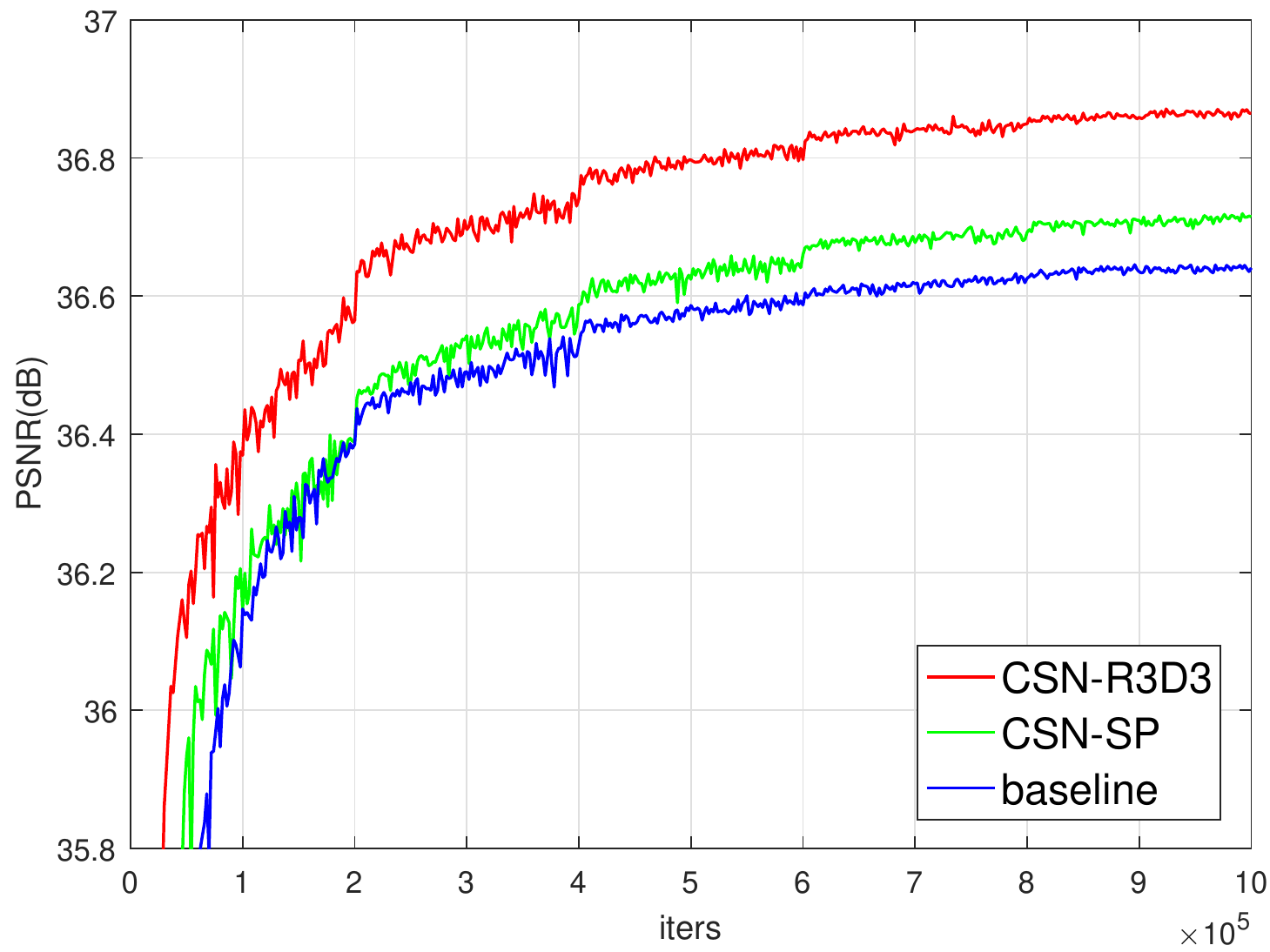}
  \end{minipage}}
  \subfigure[]{\label{fig6b}
  \begin{minipage}[t]{0.32\textwidth}
    \centering
    \includegraphics[scale = 0.38]{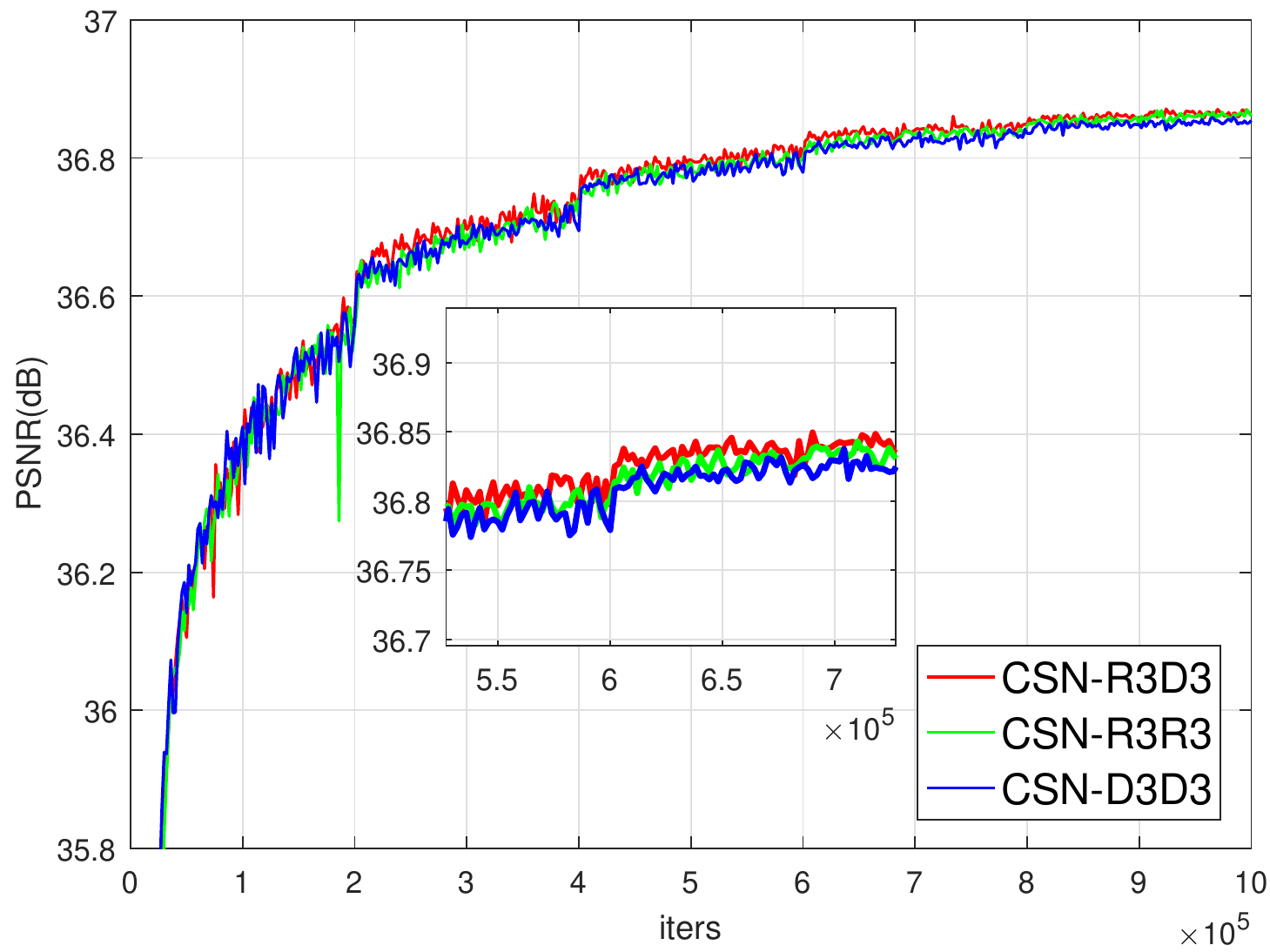}
  \end{minipage}}
  \subfigure[]{\label{fig6c}
  \begin{minipage}[t]{0.32\textwidth}
    \centering
    \includegraphics[scale = 0.38]{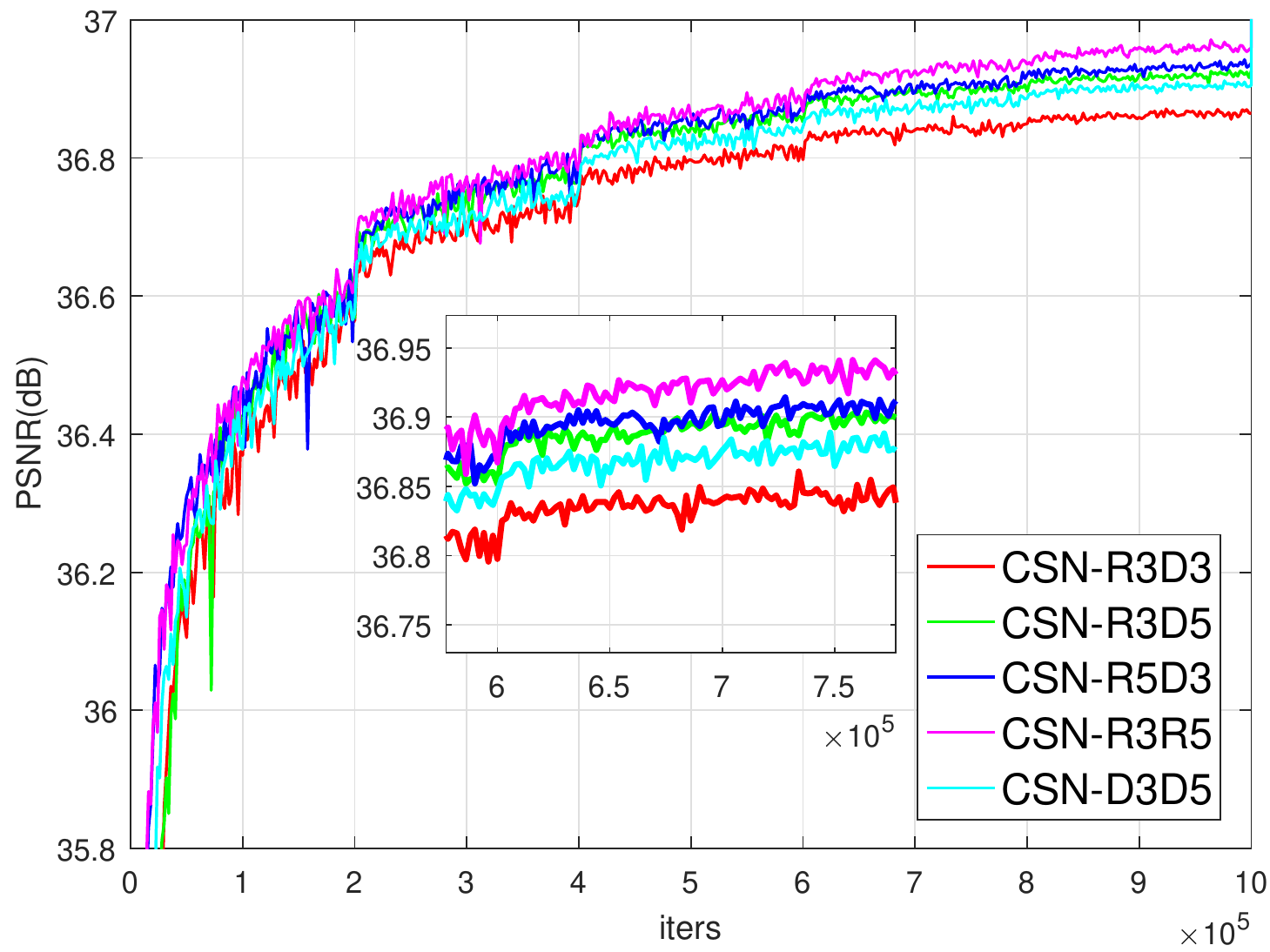}
  \end{minipage}}
  \vspace{-2mm}
  \caption{The performance comparison between different structures shown in Fig.\ref{fig5} on $\mathcal{V}(\text{T1}, \text{TD})$ for SR$\times$2 (bicubic = 31.72dB). (a) Channel splitting and the merge-and-run mapping. (b) Different branch structures. (c) Different kernel sizes. The baseline stage mapping is a single convolutional layer followed by a ReLU operation (Fig.\ref{fig5}(a)), which emphasizes the impact of channel splitting.}
  \label{fig6}
\end{figure*}

\subsubsection{Channel Splitting and Merging}
As for the $i$-th CSB, the input tensor $\mathbf{x}_{i-1}$ is first equally split into two tensors along the channel direction, $\mathbf{x}_{i-1}^{\bot}$ and $\mathbf{x}_{i-1}^{\top}$, which are the inputs of the lower (dense) branch and the upper (residual) branch respectively. It could be formally expressed as:
\begin{equation}
  \label{eqn7}
  \mathbf{x}_{i-1}^{\bot},\ \mathbf{x}_{i-1}^{\top} = \mathcal{S}_{\text{c}}(\mathbf{x}_{i-1}),\ \ \ \ i = 1,\ 2,\ \ldots,\ n,
\end{equation}
where $\mathcal{S}_{\text{c}}(\cdot)$ is the channel splitting function. It can be viewed as a unary operator that splits the input tensors into two parts along the channel direction. Through this channel splitting operation, we can apply different information transmission mechanisms on each branch. For example, we adopt residual learning and dense learning on the upper and lower branches respectively. Correspondingly, there exists a channel merging operation, $\mathcal{M}_{\text{c}}(\cdot)$, at the end of each CSB:
\begin{equation}
  \label{eqn8}
  \mathbf{x}_{i-1} = \mathcal{M}_{\text{c}}(\mathbf{x}_{i-1}^{\bot},\ \mathbf{x}_{i-1}^{\top}),\ \ i = 1,\ 2,\ \ldots,\ n.
\end{equation}
It should be noted that $\mathcal{M}_{\text{c}}(\cdot)$ is a bivariate function, while the $[\ldots]$ in (\ref{eqn4}) and (\ref{eqn5}) is a multivariate operator. The channel splitting and merging operations are expected to artificially interfere with the information flow within the network, which helps the model to process the hierarchy features with different properties in a targeted way. In addition, it is also an effective manner to maintain the scale of model parameters and increase the depth of the network.

\subsubsection{Feature Reuse and New Feature Exploration}
A CSB module assembles two residual-like branches in parallel with a merge-and-run mapping \cite{Zhao2017Deep} but each branch has different structures. Suppose the $i$-th CSB contains $m$ stage mappings and each branch in a stage includes two convolutional layers with a ReLU operation in the middle (Fig.\ref{fig:overall_structure_CSN}(b)). Denote $\mathcal{H}_{i,j}^{\bot}(\cdot)$ and $\mathcal{H}_{i,j}^{\top}(\cdot)$ as the transition functions of the lower and the upper residual branches in the $j$-th stage mapping respectively. Then the transition function of the $j$-th stage mapping can be represented in matrix form as below:
\begin{equation}
  \label{eqn9}
  \begin{bmatrix}
    \mathbf{x}_{i-1,j}^{\top} \\
    \mathbf{x}_{i-1,j}^{\bot}
  \end{bmatrix} =
  \begin{bmatrix}
    \mathcal{H}_{i,j}^{\top}(\mathbf{x}_{i-1,j-1}^{\top}) \\
    \mathcal{H}_{i,j}^{\bot}(\mathbf{x}_{i-1,j-1}^{\bot})
  \end{bmatrix} + \frac{1}{2}
  \begin{bmatrix}
    \mathbf{I} & \mathbf{I} \\
    \mathbf{I} & \mathbf{I}
  \end{bmatrix}
  \begin{bmatrix}
    \mathbf{x}_{i-1,j-1}^{\top} \\
    \mathbf{x}_{i-1,j-1}^{\bot}
  \end{bmatrix},
\end{equation}
where $i = 1,2,\ldots,n$ and $j = 1,2,\ldots,m$ are the index of the $i$-th CSB and the $j$-th stage mapping in this CSB. $\mathbf{x}_{i-1,j-1}^{\bot}$ and $\mathbf{x}_{i-1,j-1}^{\top}$ ($\mathbf{x}_{i-1,j}^{\bot}$ and $\mathbf{x}_{i-1,j}^{\top}$) are the inputs (outputs) of the $j$-th stage mapping in the $i$-th CSB, and $\mathbf{x}_{i-1,0}^{\bot} = \mathbf{x}_{i-1}^{\bot}$ and $\mathbf{x}_{i-1,0}^{\top} = \mathbf{x}_{i-1}^{\top}$ are the input tensors of the lower and upper branches respectively. $\mathbf{I}$ denotes identity matrix. Therefore, the coefficient matrix
\begin{equation}
  \mathbf{C} = \frac{1}{2}
  \begin{bmatrix}
    \mathbf{I} & \mathbf{I} \\
    \mathbf{I} & \mathbf{I}
  \end{bmatrix},
\end{equation}
is an idempotent transformation matrix of the merge-and-run mapping \cite{Zhao2017Deep,Hu2018Single}. The idempotent property can promote the information flow across the different modules and encourage gradient back-propagation during model training, similar to identity mapping \cite{He2015Identity}. It is worth noting that the upper branch is a residual-like structure similar to EDSR \cite{Lim2017Enhanced} and the lower branch is a simplified dense-like structure similar to DenseNet \cite{Huang2016Densely} or RDB \cite{Zhang2018Residual}, which uses only one skip dense connection to explore new features. Through merge-and-run mapping, we can effectively integrate the superiority of feature reuse and new feature exploration provided by the residual branch and the dense branch.

\subsubsection{Local Residual Learning (LRL)}
The feature maps from these two branches, $\mathbf{x}_{m}^{\bot}$ and $\mathbf{x}_{m}^{\top}$, are merged together after $m$ stages of merge-and-run mappings. Next, a local residual learning (LRL) \cite{Zhang2018Residual} is also introduced in the CSB to further improve the information flow. The output of this CSB module is thus given by:
\begin{equation}
  \label{eqn10}
  \mathbf{x}_{i} = \mathcal{L}(\mathcal{M}_{\text{c}}(\mathbf{x}_{i-1,m}^{\bot},\ \mathbf{x}_{i-1,m}^{\top})) + \mathbf{x}_{i-1},
\end{equation}
where $\mathcal{L}(\cdot)$ corresponds to a 1$\times$1 convolutional operation at the end of the CSB, as shown in Fig.\ref{fig:overall_structure_CSN}(b). Unlike \cite{Zhang2018Residual}, the local residual features are derived from our CSB module, instead of the densely connected block \cite{Huang2016Densely}.

\begin{table*}[t]
  \centering
  \caption{The detailed configuration of the proposed CSN model. Only one single stage mapping is shown in NMN due to the exactly same structure of each CSB and each stage mapping. All conv layers are padded to keep the size of feature maps unchanged.}
  \vspace{-2mm}
  \begin{tabular}{C{0.9cm}||C{0.75cm}|C{0.75cm}|C{0.75cm}||C{0.8cm}|C{0.8cm}|C{0.8cm}|C{0.8cm}|C{0.8cm}||C{0.8cm}|C{0.8cm}|C{1.0cm}|C{1.0cm}|C{1.0cm}}
    \toprule
    \multirow{2}{*}{Config} & \multicolumn{3}{c||}{FEN (three conv layers)} & \multicolumn{5}{c||}{NMN ($\text{stage mapping} \times m = $ CSB, $\text{CSB} \times n$)}& \multicolumn{5}{c}{IRN (Upscaling depends on the scaling factor $r$)}\\
    \cmidrule{2-14}
    & \multicolumn{3}{c||}{General convolution} & \multicolumn{2}{c|}{Residual branch} & \multicolumn{2}{c|}{Dense branch} & Merge & \multicolumn{2}{c|}{Feature Fusion} & \multicolumn{2}{c|}{Upscale ($r = 2|3,4$)} & Recover \\
    \hhline{-||---||-----||-----}
    Layer  & conv1 & conv2 & conv3 & conv1 & conv2 & conv1 & conv2 & conv1 & conv1 & conv2 &  conv1  &  conv1,2  &  conv1   \\
    \hhline{-||---|-----|-----}
    Filter & 256   & 256   & 256   & 128   & 128   & 64   & 128   & 256   & 256   & 256   & 256$\cdot r^2$ & 256$\cdot 2^2$ &  1$|$96  \\
    \hhline{-||---||-----||-----}
    Kernel & 3$\times$3 & 1$\times$1 & 3$\times$3 & 3$\times$3 & 3$\times$3 & 3$\times$3 & 3$\times$3 & 1$\times$1 & 1$\times$1 & 3$\times$3 & 3$\times$3 & 3$\times$3 & 3$\times$3  \\
    \hhline{-||---||-----||-----}
    Act & / & / & / & ReLU  &  /  &  ReLU  &  /  &  /  &  /  &  /  &  /  &  /  &  / \\
    \bottomrule
  \end{tabular}
  \label{tab0}
\end{table*}

\subsection{Multilevel Residual Mechanism}
Normally, the LR images and the corresponding HR images share same information to a large extent, which indicates that a large
part of the topological structure of their high-dimensional manifolds are similar to each other. Therefore, it is beneficial to explicitly allow the model to learn the residual between the \textit{original} LR input and the HR output \cite{Kim2016Accurate}. However, because LR and HR images have different sizes, the residual between them cannot be directly obtained. We adopt a bicubic interpolated version of the LR image to match the size of the HR image, and use it to approximate the residual between the original LR image and the HR image. This is implemented by simply adding the interpolated image to the output of the last convolutional layer of the entire network, which we term as external skip connection (ESC) (Fig.\ref{fig:overall_structure_CSN}(a)). Thus, (\ref{eqn6}) should be rewritten as:
\begin{equation}
  \label{eqn11}
  \mathbf{y} = \mathcal{F}_{\text{R}}(\mathbf{x}_{\text{R}}) + \hat{\mathbf{x}},
\end{equation}
where $\hat{\mathbf{x}}$ is the interpolated version of the original LR input $\mathbf{x}$. Although we use bicubic interpolation here, one can use any other interpolation algorithm (e.g., nearest neighbour, bilinear and B-Spline etc.).

Combined with GSC and LRL, the whole network shows a characteristic of multilevel residual learning. Our experience shows that this can further stabilize the training process, and even help to slightly improve the model performance. Because the degeneration of MR training samples causes model training more unstable, it is especially helpful for the task of single MR image super-resolution.

\begin{table}[t]
  \centering
  \caption{The testing performance of different structures shown in Fig.\ref{fig5} on $\mathcal{T}(\text{T1}, \text{TD})$ for SR$\times$2. The maximal values of each column are in {\textcolor[rgb]{1,0,0}{red}}, and the second ones are in {\textcolor[rgb]{0,0,1}{blue}}.}
  \vspace{-2mm}
  \begin{tabular}{C{1.6cm}|C{1.3cm}|C{1.3cm}|C{1.3cm}|C{1.1cm}}
    \toprule
    Network       & Network    & Network     & \multirow{2}{*}{PSNR (dB)} &  \multirow{2}{*}{SSIM} \\
    Configuration & Parameters & Depth       &                            &         \\
    \hline
    Bicubic       & /                 & /  & 33.38          & 0.9460  \\
    \hline
    baseline      & 13643521          & 27 & 38.37          & 0.9803 \\
    CSN-SP        & 13646593          & 43 & 38.46          & 0.9807 \\
    CSN-R3D3      & 13646593          & 43 & 38.62          & 0.9813 \\
    \hline
    CSN-R3R3      & 13647614          & 43 & 38.61          & 0.9813 \\
    CSN-D3D3      & 13645566          & 43 & 38.59          & 0.9812 \\
    \hline
    CSN-R3D5      & 22035198          & 43 & 38.67          & 0.9815 \\
    CSN-R5D3      & {\textcolor[rgb]{0,0,1}{22035198}} & {\textcolor[rgb]{0,0,1}{43}} & {\textcolor[rgb]{0,0,1}{38.68}} & {\textcolor[rgb]{0,0,1}{0.9816}} \\
    CSN-R3R5      & {\textcolor[rgb]{1,0,0}{22036225}} & {\textcolor[rgb]{1,0,0}{43}} & {\textcolor[rgb]{1,0,0}{38.70}} & {\textcolor[rgb]{1,0,0}{0.9817}} \\
    CSN-D3D5      & 22034177          & 43 & 38.64          & 0.9814 \\
    \bottomrule
  \end{tabular}
  \label{tab1}
\end{table}

\begin{table}[t]
  \centering
  \caption{Test results of the models with different ESC approximations on $\mathcal{T}$(PD, BD). The maximal PSNR and SSIM values of each row are in {\textcolor[rgb]{1,0,0}{red}}, and the second ones are in {\textcolor[rgb]{0,0,1}{blue}}.}
  \vspace{-2mm}
  \begin{tabular}{C{0.6cm}|C{1.5cm}|C{1.5cm}|C{1.5cm}|C{1.5cm}}
    \toprule
    \multirow{2}{*}{scale}  & ESC-          & ESC-          & ESC-          & ESC-    \\
                            & None          & NN            & Bilinear     & Bicubic \\
    \hline
    $\times$2             & {\textcolor[rgb]{0,0,1}{41.20/0.9893}} & 41.19/0.9893 & 41.18/0.9893 & {\textcolor[rgb]{1,0,0}{41.28/0.9895}} \\
    $\times$3             & 35.80/0.9688 & 35.79/0.9688 & {\textcolor[rgb]{0,0,1}{35.82/0.9689}} & {\textcolor[rgb]{1,0,0}{35.87/0.9693}} \\
    $\times$4             & 33.32/0.9478 & 33.31/0.9482 & {\textcolor[rgb]{0,0,1}{33.34/0.9483}} & {\textcolor[rgb]{1,0,0}{33.40/0.9486}} \\
    \bottomrule
  \end{tabular}
  \label{tab2}
\end{table}

\subsection{Training Objective and Network Depth}
Our model is a typical end-to-end mapping from LR images to HR images. The estimation of model parameters is achieved by minimizing the loss between the reconstructed HR images and the ground truth HR images. Given a training dataset $\mathcal{D} = \{\mathbf{x}^{(i)}, \mathbf{y}^{(i)}\},\ i=1,2,\ldots,|\mathcal{D}|$, where $|\mathcal{D}|$ is the total number of training samples, we use $l_1$ loss for model training:
\begin{equation}
\label{12}
L(\boldsymbol{\theta}) = \frac{1}{|\mathcal{D}|}\sum_{i = 1}^{|\mathcal{D}|}||\mathbf{y}^{(i)} - \mathcal{F}_{\text{CSN}}(\mathbf{x}^{(i)}; \boldsymbol{\theta})||_{1},
\end{equation}
where $\boldsymbol{\theta}$ indicates the set of model parameters, and $\mathcal{F}_{\text{CSN}}(\cdot)$ is the mapping function of the entire CSN model. $\mathbf{y}^{(i)}$ is the HR target corresponding to the LR input $\mathbf{x}^{(i)}$. Despite that minimizing ${l}_{2}$ loss is generally preferred since it maximizes the peak signal to noise ratio (PSNR), ${l}_{1}$ loss provides better convergence for model training \cite{Lim2017Enhanced}. This is especially helpful in case of the degradation of training samples.

The depth of a deep network is usually defined as the longest path from the input to the output. Thus, the depth of the overall CSN model is given by:
\begin{equation}
  \label{eqn13}
  D = n(2m + 1) + s + 6,
\end{equation}
where $n$ is the number of CSBs in the entire network and $m$ is the number of stage mappings in each CSB. $s$ represents the depth of the pixel shuffle layer. Note that $s$ depends on the scaling factor \cite{Lim2017Enhanced}, i.e., $s=1$ for SR$\times$2 and SR$\times$3, and $s=2$ for SR$\times$4.

\begin{figure}[t]
  \centering
  \includegraphics[width = 0.4\textwidth]{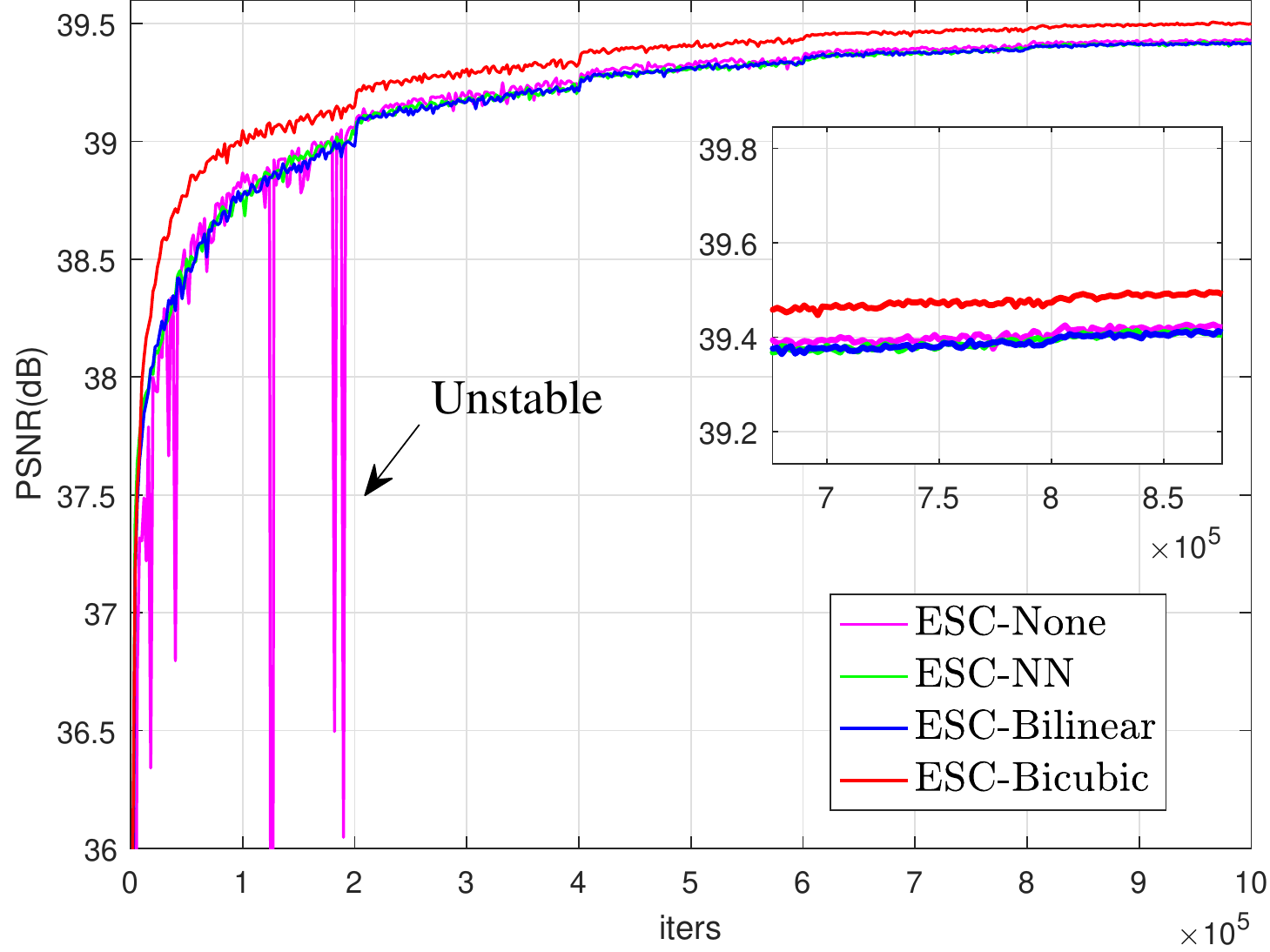}   \\
  \vspace{-2mm}
  \caption{The impact of different ESC approximations on model training and performance. The comparison is carried out on $\mathcal{D}$(PD, BD) for SR$\times$2. Note that ESC-None shows obvious instability.}\label{fig7}
\end{figure}

\begin{figure*}[t]
  \centering
  \subfigure[Impact of building blocks ($n$)]{\label{fig8a}
  \begin{minipage}[t]{0.32\textwidth}
    \centering
    \includegraphics[scale = 0.38]{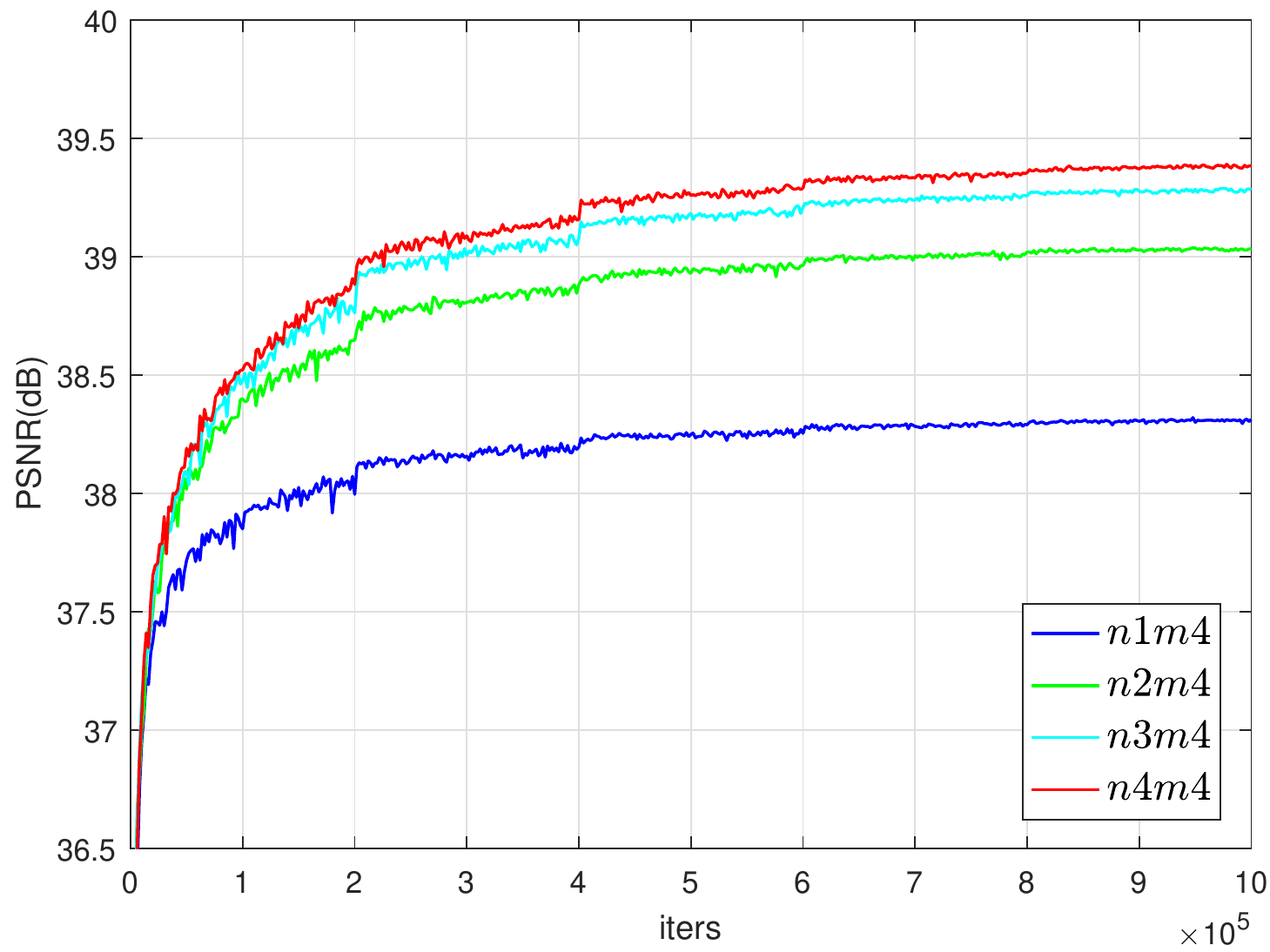}
  \end{minipage}}
  \subfigure[Impact of stage mappings ($m$)]{\label{fig8b}
  \begin{minipage}[t]{0.32\textwidth}
    \centering
    \includegraphics[scale = 0.38]{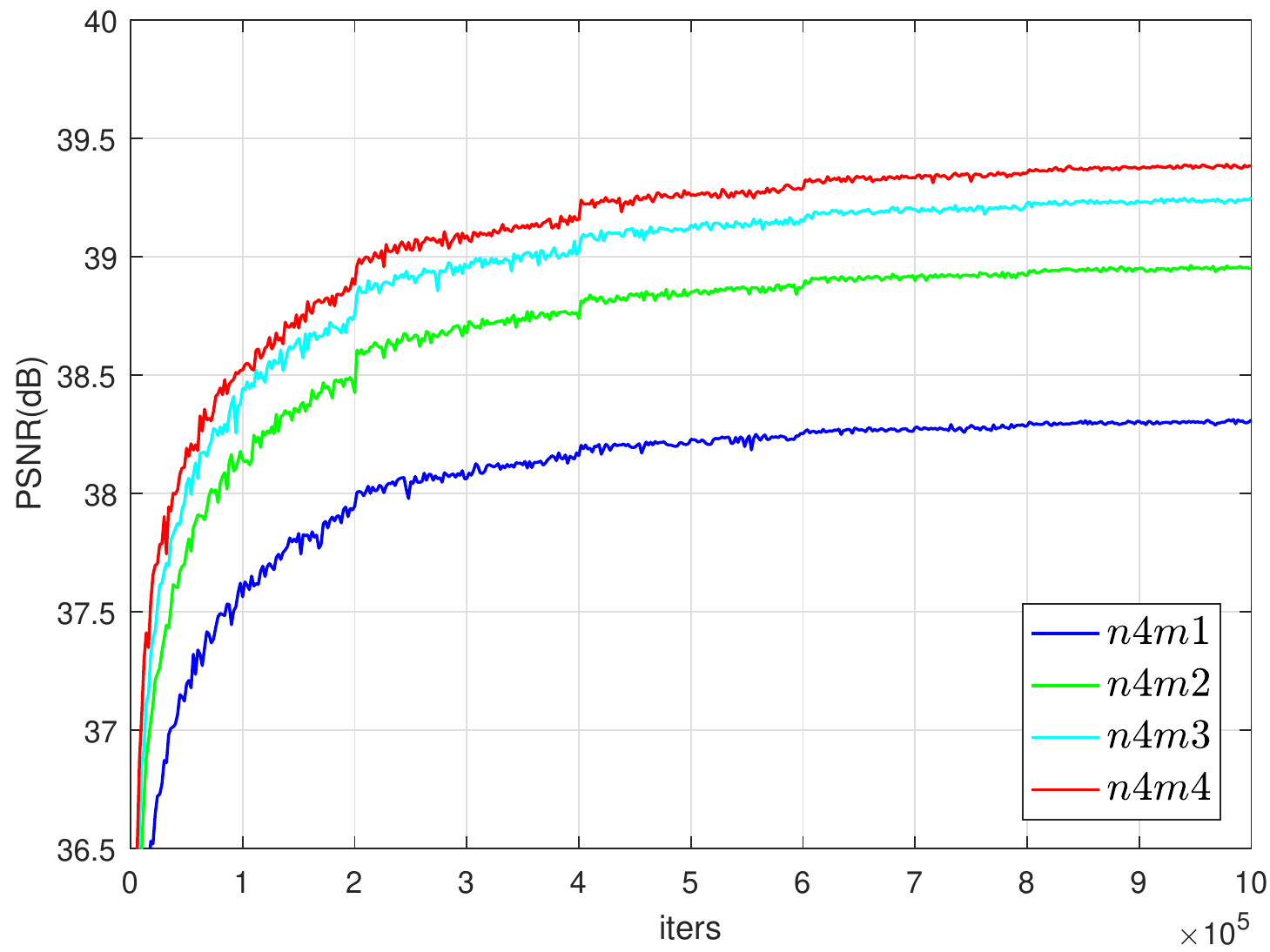}
  \end{minipage}}
  \subfigure[Comparison on $\mathcal{T}(\text{T2}, \text{TD})$]{\label{fig8c}
  \begin{minipage}[t]{0.32\textwidth}
    \centering
    \includegraphics[scale = 0.38]{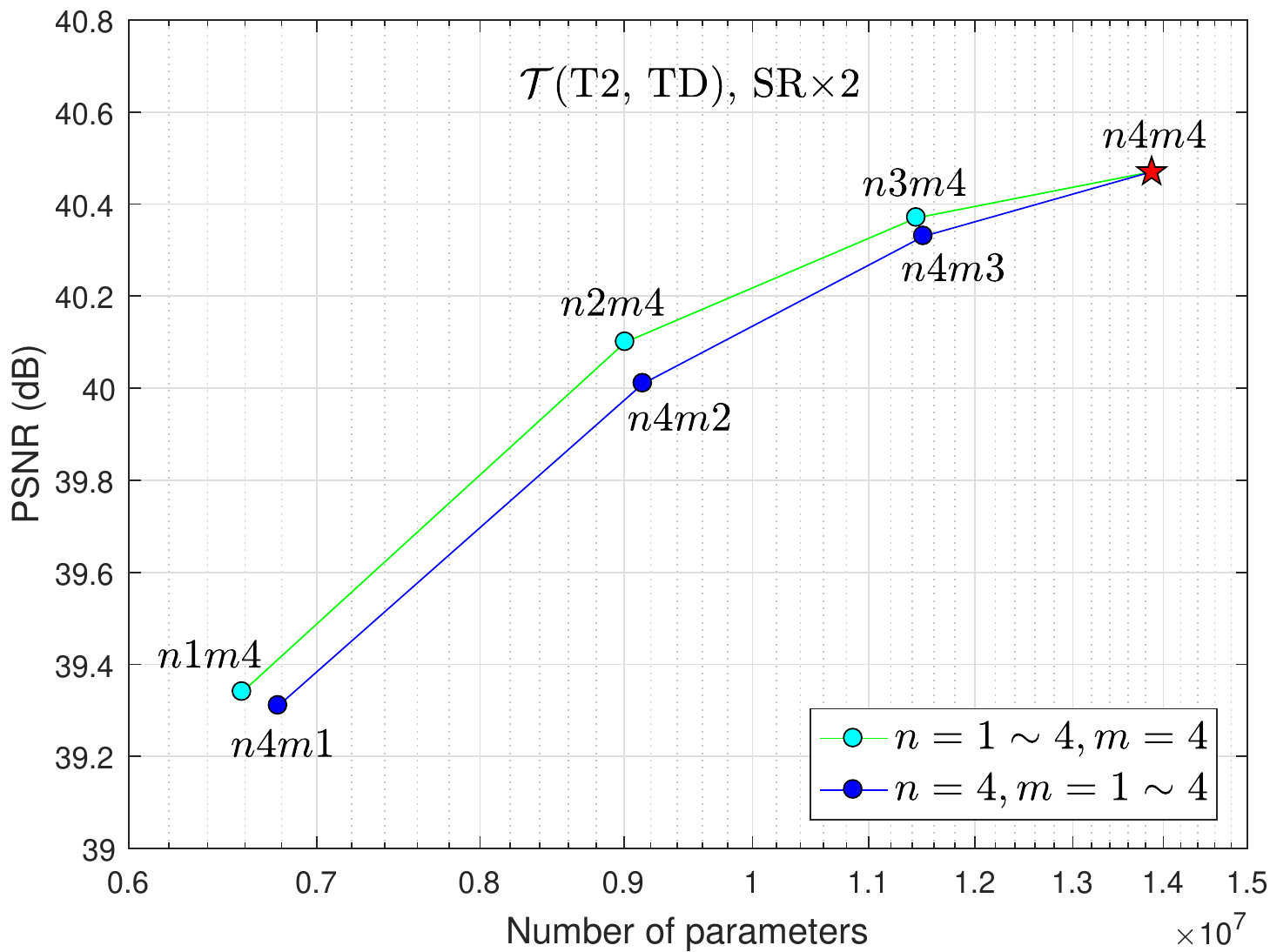}
  \end{minipage}}
  \vspace{-2mm}
  \caption{The performance comparison between the models with different number of stage mappings and building blocks. (a) and (b): The validation performance of the models on $\mathcal{V}(\text{T2}, \text{TD})$ (SR$\times$2, Bicubic: 31.92dB). (c) The testing performance of all compared models on $\mathcal{T}(\text{T2}, \text{TD})$ (SR$\times$2, Bicubic: 33.06dB).}
  \label{fig8}
\end{figure*}

\section{Experimental Results}
\label{sec:experiments}

In this section, we first introduce the generation of training examples and the implementation details. Then we investigate the impact of different configurations of CSB and the whole network on SR performance. Next, our CSN model is compared with several typical SISR methods under two common image degradations: bicubic downsampling (BD) and $k$-space truncation (TD). We use PSNR and structural similarity index metric (SSIM) \cite{Wang2004Image} as the metrics of quantitative evaluation.

\subsection{Dataset and Sample Generation}
The IXI dataset is used to construct our SR datasets, and it contains three types of MR images: 581 T1 volumes, 578 T2 volumes and 578 PD volumes. Firstly, we take the intersection of these three subsets, resulting in 576 3D volumes for each type of MR images. These 3D volumes are then clipped to the size of 240$\times$240$\times$96 (height$\times$width$\times$depth) to fit 3 scaling factors ($\times$2, $\times$3 and $\times$4). In this work, we only focus on the in-plane SR of 2D MR slices. Therefore, each 3D MR volume contains 96 training samples with a single channel. The LR images are generated according to bicubic downsampling and $k$-space truncation. As for truncation degradation, the HR images are first converted into $k$-space by discrete Fourier transform (DFT) and then truncated along both height and width directions (Fig.\ref{fig4}). We randomly selected 500 volumes for training ($\mathcal{D}$), 70 volumes for testing ($\mathcal{T}$) and the remaining for quick validation ($\mathcal{V}$). We employ the convention: \textit{dataset name (MR image type, image degradation model)}, to indicate a specific sub dataset for convenience. For instance, $\mathcal{D}(\text{T2}, \text{BD})$ represents the T2 training dataset with bicubic degradation and $\mathcal{T}(\text{PD}, \text{TD})$ represents the PD testing dataset with $k$-space truncation degradation. The processed datasets are available at: \url{https://pan.baidu.com/s/1Ak3GiJk5H1Pdn3igzElb7w} (\textbf{kn3d}).

At present, the model only targets at the task of single 2D MR image super-resolution. Thus, we have $500 \times 96 = 48000$ training examples in a single training dataset. The generated datasets can be conveniently applied to develop 3D algorithms as each dimension is clipped to the common multiple of 2, 3 and 4, which will be a part of our future work.

\subsection{Implementation Details}
The configuration of the model is shown in Fig.\ref{fig:overall_structure_CSN} with $n=m=4$. The size of minibatch and the number of feature maps are set to 16 and 256 respectively. For the dense branch within a CSB, the growth ($g$ in Fig.\ref{fig3} and Fig.\ref{fig5}) is set to 64. If not specified, the kernel size follows the annotation of Fig.\ref{fig:overall_structure_CSN}.

We train the models by using image patches of size 24$\times$24 randomly extracted from LR slices with the corresponding HR patches. Data augmentation is simply implemented by random horizontal flips and 90$^{\circ}$ rotations, as \cite{Lim2017Enhanced} and \cite{Zhang2018Residual}. All models are implemented (or reimplemented) in TensorFlow 1.7.0 and trained on a NVIDIA GeForce GTX 1080 Ti GPU for one million iterations. We adopt Xavier initialization \cite{Glorot2010Understanding} for all model parameters and Adam optimizer \cite{Kingma2014Adam} to minimize the loss by setting ${\beta}_{1} = 0.9$, ${\beta}_{2} = 0.999$ and $\epsilon = 10^{-8}$. Learning rate is initialized as $10^{-4}$ for all layers and halved at every $2 \times 10^{5}$ iterations i.e., piecewise constant decay.

\begin{table*}[t]
  \centering
  \caption{Quantitative comparison between different methods on 6 test datasets (2 image degradations and 3 MR image types). The maximal PSNR (dB) and SSIM values of each comparison group are marked in {\textcolor[rgb]{1,0,0}{red}}, and the second ones are marked in {\textcolor[rgb]{0,0,1}{blue}} (PSNR / SSIM).}
  \vspace{-2mm}
  \begin{tabular}{R{2.1cm}|C{0.7cm}|C{0.7cm}||C{1.75cm}|C{1.75cm}|C{1.75cm}||C{1.75cm}|C{1.75cm}|C{1.75cm}}
    \toprule
    \multirow{2}{*}{method $\backslash$ dataset} &\multirow{2}{*}{mode} & \multirow{2}{*}{scale} &  \multicolumn{3}{c||}{bicubic downsampling $\mathcal{T}$(:, BD)} &  \multicolumn{3}{c}{$k$-space truncation $\mathcal{T}$(:, TD)}  \\
    \cmidrule{4-9}
    & &   &  PD &  T1 &  T2 & PD & T1 & T2 \\
    \hhline{---||---||---}
    Bicubic [2D] & \multirow{8}{*}{C1} & $\times$2 & 35.04 / 0.9664 & 33.80 / 0.9525 & 33.44 / 0.9589 & 34.65 / 0.9625 & 33.38 / 0.9460 & 33.06 / 0.9541 \\
    NLM \cite{Manjon2010Nonlocal} && $\times$2 & 37.26 / 0.9773 & 35.80 / 0.9685 & 35.58 / 0.9722 & 36.18 / 0.9707 & 34.71 / 0.9581 & 34.56 / 0.9641 \\
    SRCNN \cite{Dong2016Image}    && $\times$2 & 38.96 / 0.9836 & 37.12 / 0.9761 & 37.32 / 0.9796 & 38.23 / 0.9802 & 36.52 / 0.9705 & 37.04 / 0.9773 \\
    VDSR \cite{Kim2016Accurate}   && $\times$2 & 39.97 / 0.9861 & 37.67 / 0.9783 & 38.65 / 0.9836 & 39.89 / 0.9850 & 37.58 / 0.9760 & 38.74 / 0.9823 \\
    RDN \cite{Zhang2018Residual}  && $\times$2 & 40.31 / 0.9870 & 37.95 / 0.9795 & 38.75 / 0.9838 & 40.39 / 0.9862 & 38.08 / 0.9784 & {\textcolor[rgb]{0,0,1}{40.02}} / 0.9826 \\
    CMSCN \cite{Hu2018Single}     && $\times$2 & {\textcolor[rgb]{0,0,1}{40.84 / 0.9883}} & {\textcolor[rgb]{0,0,1}{38.06 / 0.9800}} & {\textcolor[rgb]{0,0,1}{39.54 / 0.9857}} & {\textcolor[rgb]{0,0,1}{41.14 / 0.9882}} & {\textcolor[rgb]{0,0,1}{38.23 / 0.9795}} & 39.63 / 0.9845 \\
    FSCWRN \cite{Shi2019MR}       && $\times$2 & 40.72 / 0.9880 & 37.98 / 0.9797 & 39.44 / 0.9855 & 40.91 / 0.9876 & 38.04 / 0.9786 & 39.82 / {\textcolor[rgb]{0,0,1}{0.9851}} \\
    CSN [Ours]                    && $\times$2 & {\textcolor[rgb]{1,0,0}{41.28 / 0.9895}} & {\textcolor[rgb]{1,0,0}{38.27 / 0.9810}} & {\textcolor[rgb]{1,0,0}{39.71 / 0.9863}} & {\textcolor[rgb]{1,0,0}{41.77 / 0.9897}} & {\textcolor[rgb]{1,0,0}{38.62 / 0.9813}} & {\textcolor[rgb]{1,0,0}{40.47 / 0.9868}} \\
    \hhline{---||---||---}
    EDSR \cite{Lim2017Enhanced} & \multirow{2}{*}{C96} & $\times$2 & 39.87 / 0.9857 & 37.56 / 0.9774 & 38.28 / 0.9824 & 39.47 / 0.9837 & 37.09 / 0.9741 & 38.11 / 0.9803 \\
    CSN [Ours]                  && $\times$2 & 40.15 / 0.9865 & 37.60 / 0.9778 & 38.53 / 0.9831 & 39.50 / 0.9839 & 36.99 / 0.9737 & 38.20 / 0.9807 \\
    \bottomrule 
    Bicubic [2D] & \multirow{8}{*}{C1} & $\times$3 & 31.20 / 0.9230 & 30.15 / 0.8900 & 29.80 / 0.9093 & 30.88 / 0.9167 & 29.79 / 0.8793 & 29.50 / 0.9016 \\
    NLM \cite{Manjon2010Nonlocal} && $\times$3 & 32.81 / 0.9436 & 31.74 / 0.9216 & 31.28 / 0.9330 & 32.02 / 0.9324 & 30.83 / 0.9027 & 30.57 / 0.9197 \\
    SRCNN \cite{Dong2016Image}    && $\times$3 & 33.60 / 0.9516 & 32.17 / 0.9276 & 32.20 / 0.9440 & 32.90 / 0.9432 & 31.72 / 0.9187 & 31.80 / 0.9381 \\
    VDSR \cite{Kim2016Accurate}  && $\times$3 & 34.66 / 0.9599 & 32.91 / 0.9378 & 33.47 / 0.9559 & 34.27 / 0.9555 & 32.57 / 0.9304 & 33.23 / 0.9515 \\
    RDN \cite{Zhang2018Residual} && $\times$3 & 35.08 / 0.9628 & {\textcolor[rgb]{0,0,1}{33.31 / 0.9430}} & 33.91 / 0.9591 & 35.00 / 0.9609 & {\textcolor[rgb]{0,0,1}{33.33 / 0.9416}} & 33.99 / 0.9576 \\
    CMSCN \cite{Hu2018Single}    && $\times$3 & 35.26 / 0.9641 & 33.25 / 0.9423 & 34.16 / 0.9613 & {\textcolor[rgb]{0,0,1}{35.41 / 0.9638}} & 33.18 / 0.9398 & {\textcolor[rgb]{0,0,1}{34.45 / 0.9611}} \\
    FSCWRN \cite{Shi2019MR}      && $\times$3 & {\textcolor[rgb]{0,0,1}{35.37 / 0.9653}} & 33.24 / 0.9423 & {\textcolor[rgb]{0,0,1}{34.27 / 0.9618}} & 35.30 / 0.9636 & 33.09 / 0.9390 & 34.34 / 0.9603 \\
    CSN [Ours]                   && $\times$3 & {\textcolor[rgb]{1,0,0}{35.87 / 0.9693}} & {\textcolor[rgb]{1,0,0}{33.53 / 0.9464}} & {\textcolor[rgb]{1,0,0}{34.64 / 0.9647}} & {\textcolor[rgb]{1,0,0}{36.09 / 0.9697}} & {\textcolor[rgb]{1,0,0}{33.68 / 0.9464}} & {\textcolor[rgb]{1,0,0}{34.95 / 0.9653}} \\
    \hhline{---||---||---}
    EDSR \cite{Lim2017Enhanced} & \multirow{2}{*}{C96} & $\times$3 & 34.39 / 0.9578 & 32.76 / 0.9347 & 33.15 / 0.9528 & 33.97 / 0.9531 & 32.27 / 0.9274 & 32.89 / 0.9482 \\
    CSN [Ours]                && $\times$3 & 34.68 / 0.9598 & 32.83 / 0.9360 & 33.36 / 0.9547 & 34.12 / 0.9540 & 32.25 / 0.9266 & 33.00 / 0.9490 \\
    \bottomrule 
    Bicubic [2D] & \multirow{8}{*}{C1} & $\times$4 & 29.13 / 0.8799 & 28.28 / 0.8312 & 27.86 / 0.8611 & 28.82 / 0.8713 & 27.96 / 0.8182 & 27.60 / 0.8511 \\
    NLM \cite{Manjon2010Nonlocal} && $\times$4 & 30.27 / 0.9044 & 29.31 / 0.8655 & 28.85 / 0.8875 & 29.27 / 0.8906 & 28.68 / 0.8439 & 28.37 / 0.8718 \\
    SRCNN \cite{Dong2016Image}    && $\times$4 & 31.10 / 0.9181 & 29.90 / 0.8796 & 29.69 / 0.9052 & 30.52 / 0.9078 & 29.31 / 0.8616 & 29.32 / 0.8960 \\
    VDSR \cite{Kim2016Accurate}   && $\times$4 & 32.09 / 0.9311 & 30.57 / 0.8932 & 30.79 / 0.9240 & 31.69 / 0.9244 & 30.14 / 0.8818 & 30.51 / 0.9162 \\
    RDN \cite{Zhang2018Residual}  && $\times$4 & 32.73 / 0.9387 & {\textcolor[rgb]{0,0,1}{31.05 / 0.9042}} & 31.45 / 0.9324 & 32.64 / 0.9362 & {\textcolor[rgb]{0,0,1}{31.00 / 0.9018}} & 31.49 / 0.9301 \\
    CMSCN \cite{Hu2018Single}     && $\times$4 & 32.53 / 0.9374 & 30.83 / 0.8997 & 31.32 / 0.9312 & 32.23 / 0.9321 & 30.55 / 0.8920 & 31.28 / 0.9278 \\
    FSCWRN \cite{Shi2019MR}       && $\times$4 & {\textcolor[rgb]{0,0,1}{32.91 / 0.9415}} & 30.96 / 0.9022 & {\textcolor[rgb]{0,0,1}{31.71 / 0.9359}} & {\textcolor[rgb]{0,0,1}{32.78 / 0.9387}} & 30.79 / 0.8973 & {\textcolor[rgb]{0,0,1}{31.71 / 0.9334}} \\
    CSN [Ours]                    && $\times$4 & {\textcolor[rgb]{1,0,0}{33.40 / 0.9486}} & {\textcolor[rgb]{1,0,0}{31.23 / 0.9093}} & {\textcolor[rgb]{1,0,0}{32.05 / 0.9413}} & {\textcolor[rgb]{1,0,0}{33.51 / 0.9489}} & {\textcolor[rgb]{1,0,0}{31.27 / 0.9092}} & {\textcolor[rgb]{1,0,0}{32.28 / 0.9421}} \\
    \hhline{---||---||---}
    EDSR \cite{Lim2017Enhanced} & \multirow{2}{*}{C96} & $\times$4 & 31.80 / 0.9284 & 30.46 / 0.8902 & 30.52 / 0.9198 & 31.44 / 0.9219 & 30.04 / 0.8803 & 30.31 / 0.9137 \\
    CSN [Ours]                && $\times$4 & 32.19 / 0.9325 & 30.53 / 0.8915 & 30.81 / 0.9231 & 31.72 / 0.9246 & 30.07 / 0.8794 & 30.54 / 0.9163 \\
    \bottomrule
  \end{tabular}
  \label{tab:C1}
\end{table*}

\subsection{Model Analysis}
In this section, we study several components of the proposed model, including the structure of stage mapping, multilevel residual learning, global feature fusion and building block utilization. The structure of the entire network and the building block refers to Fig.\ref{fig:overall_structure_CSN}.

\subsubsection{Channel Splitting Block}
The proposed stage mapping can be configured in several ways, thus equipping different CSB modules. For comparison, we have studied the structure of different stage mappings from the following aspects:
\begin{itemize}
  \item[$\ast$] If without channel splitting, four convolutional layers in a stage mapping correspond to a single convolutional layer with nearly the same number of model parameters. It is a reference structure of a stage mapping and we take it as the baseline, as shown in Fig.\ref{fig5}(a).

  \item[$\ast$] To investigate the role of the MAR mapping, we remove it from the proposed CSB and obtain the structure shown in Fig.\ref{fig5}(e). We term it as CSN-SP, where S means splitting and P means plain.

  \item[$\ast$] We also design two stage mappings shown in Fig.\ref{fig5}(b) and Fig.\ref{fig5}(f) to check the effect of different branch structures on the performance of the model. They are referred as CSN-R3R3 and CSN-D3D3 respectively, where R and D represent residual branch and dense branch, and the numbers indicate the kernel size.

  \item[$\ast$] To study the impact of different kernel sizes, four other structures are designed. They are termed as CSN-R3D5 (Fig.\ref{fig5}(c)), CSN-R5D3 (Fig.\ref{fig5}(g)), CSN-R3R5 (Fig.\ref{fig5}(d)) and CSN-D3D5 (Fig.\ref{fig5}(h)), respectively.
\end{itemize}
The proposed stage mapping structure (as shown in Fig.\ref{fig:overall_structure_CSN}(b) and Fig.\ref{fig3}(c)) is marked as CSN-R3D3. Both $m$ and $n$ are set to 4 for all experiments in this section.

The performance of the compared stage mapping structures on $\mathcal{V}(\text{T1}, \text{TD})$ for SR$\times$2 is shown in Fig.\ref{fig6}. It can be seen from Fig.\ref{fig6a} that both channel splitting and the MAR mapping can significantly improve the model performance. According to Fig.\ref{fig6b}, the performance of CSN-R3D3 is slightly better than that of CSN-R3R3 and CSN-D3D3. However, it is noteworthy that \textit{the model parameters R3R3$>$R3D3$>$D3D3, and the depth of these networks is the same}, implying that mixing different branch structures is indeed helpful to boost the performance of the model, although slightly. It is observed from Fig.\ref{fig6c} that the model performance R3R5$>$R5D3$>$R3D5$>$D3D5$>$R3D3. However, the parameters of the first four structures are about 1.6 times that of CSN-R3D3, causing a worse tradeoff between model performance and model scale. We can assume that their performance improvement on CSN-R3D3 is mainly due to the increase of model parameters. In addition, we can find that the residual branch favors better performance. The conclusions are further verified by the testing results shown in Table \ref{tab1}.

\subsubsection{External Skip Connection}
To investigate the impact of external skip connections (ESC), we build three other models according to our CSN model, two of which use nearest neighbor (NN) and bilinear respectively to approximate the residual between the original LR input $\mathbf{x}$ and the corresponding HR target $\mathbf{y}$, and the other does not use ESC. They are termed as ESC-None, ESC-NN, ESC-Bilinear, and the one we use is termed as ESC-Bicubic. We train these models on $\mathcal{D}$(PD, BD) and the validation performance is plotted in Fig.\ref{fig7}. It can be easily observed that ESC-Bicubic perform significantly better than other models. The corresponding results on $\mathcal{T}$(PD, BD) also illustrate this conclusion (Table \ref{tab2}).

Another important observation is that the ESC contributes to stable model training, no matter which interpolation method is used. Therefore, the ESC can reduce the possibility of training failure, which is also beneficial in the case of the degradation of training examples.

\subsubsection{The Number of Stage Mappings and Blocks}
It can be seen from (\ref{eqn13}) that the network depth $D$ is mainly determined by the number of CSBs $m$ and the number of stage mappings $n$. We examine the impact of these two hyperparameters on the performance of the model. Firstly, we fix $m$ to 4 and change $n$ from 1 to 4. Fig.\ref{fig8a} displays the evolution curves of PSNR performance on $\mathcal{V}$(T2, TD) for SR$\times$2. It can be seen that the performance is improved gradually with the increased number of building blocks, but at the expense of increased parameters. Next, we fix $n$ to 4 and change $m$ from 1 to 4. The PSNR curves of the models on the same dataset are plotted in Fig.\ref{fig8b}. We observe a similar trend of the curves as $m$ changes. The result is unsurprising because increasing $m$ or $n$ increases the network depth and model parameters.

Finally, we show the final SR performance of all compared models on the corresponding testing dataset $\mathcal{T}$(T2, TD) in Fig.\ref{fig8c}, versus the number of parameters. It is worth noting that the models with $t$ building blocks and 4 stage mappings perform better than the models with 4 building blocks and $t$ stage mappings ($t = 1,2,3$), although the former has fewer model parameters. In next experiments, we choose $n = m = 4$ for our CSN model. Therefore, the network depth is 43 for SR$\times$2 and SR$\times$3, and 44 for SR$\times$4.

\begin{figure*}[t]
  \centering
  \includegraphics[width = \textwidth]{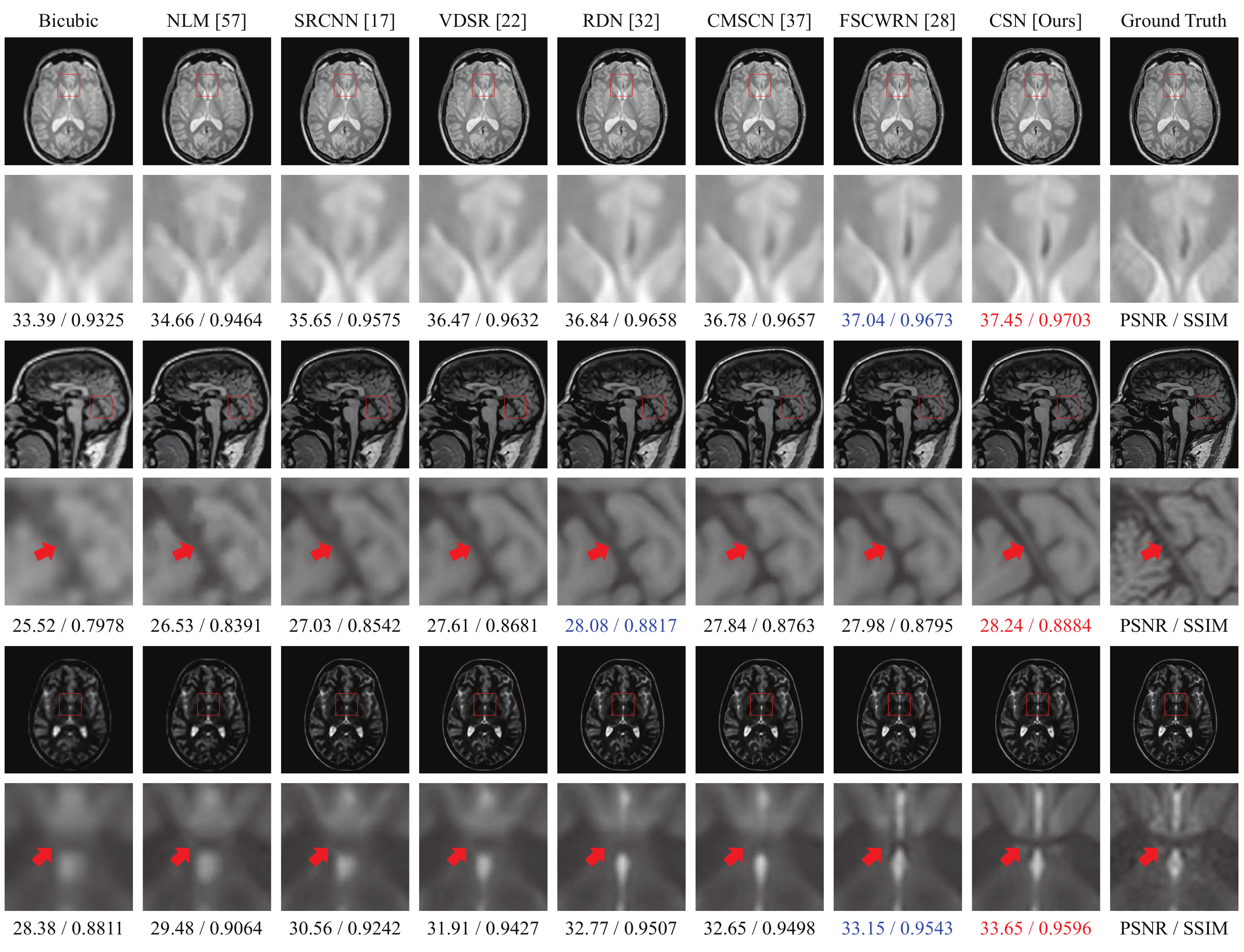}   \\
  \vspace{-2mm}
  \caption{The visual effect of the compared methods on a PD (top), T1 (middle) and T2 (bottom) image with SR$\times$3, SR$\times$4 and SR$\times$4, respectively. The image degradation is \textbf{bicubic downsampling} (C1). The maximal PSNR (dB) and SSIM values for each displayed image are in {\textcolor[rgb]{1,0,0}{red}}, and the second ones are in {\textcolor[rgb]{0,0,1}{blue}}.}\label{fig9}
\end{figure*}

\subsection{Comparison with Other Methods}
To further illustrate the effectiveness of the proposed CSN model, we compare it with several advanced SISR methods quantitatively and qualitatively, including NLM \cite{Manjon2010Nonlocal}, SRCNN \cite{Dong2016Image}, VDSR \cite{Kim2016Accurate}, RDN \cite{Zhang2018Residual}, CMSCN \cite{Hu2018Single}, FSCWRN \cite{Shi2019MR} and EDSR \cite{Lim2017Enhanced}. These models are retrained on the generated datasets with all image types and scaling factors. As mentioned earlier, the degradation of training samples may lead to training failure of some models, especially for those with extremely deep structure and large number of parameters, e.g., EDSR \cite{Lim2017Enhanced}. To solve the problem, we train the EDSR by taking 96 slices of a 3D volume as 96 channels of a 2D sample. This can effectively avoid the training failure problem but at the cost of accuracy reduction. We attach C1 and C96 to the model name to mark these two cases. For fair comparison, we train both CSN (C1) and CSN (C96).

\begin{figure*}[t]
  \centering
  \includegraphics[width = \textwidth]{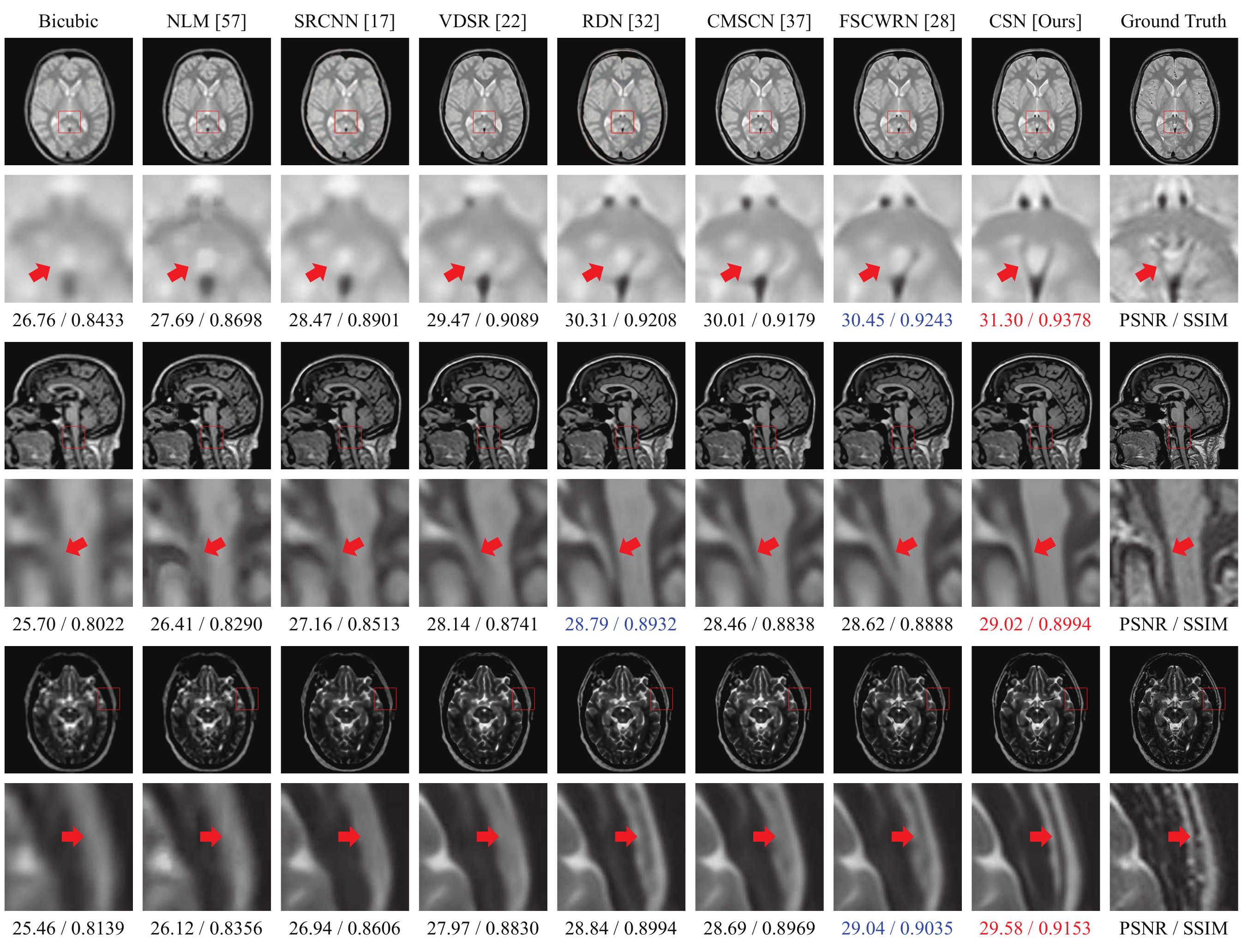}   \\
  \caption{The visual effect of the compared methods on a PD (top), T1 (middle) and T2 (bottom) image with scaling factor SR$\times$4. The image degradation is \textbf{$k$-space truncation} (C1). The maximal PSNR (dB) and SSIM values for each displayed image are in {\textcolor[rgb]{1,0,0}{red}}, and the second ones are in {\textcolor[rgb]{0,0,1}{blue}}.}\label{fig10}
\end{figure*}

\begin{figure*}[!htbp]
  \centering
  \subfigure[bicubic downsampling (BD)]{\label{fig11a}
  \begin{minipage}[t]{0.48\textwidth}
    \centering
    \includegraphics[scale = 0.68]{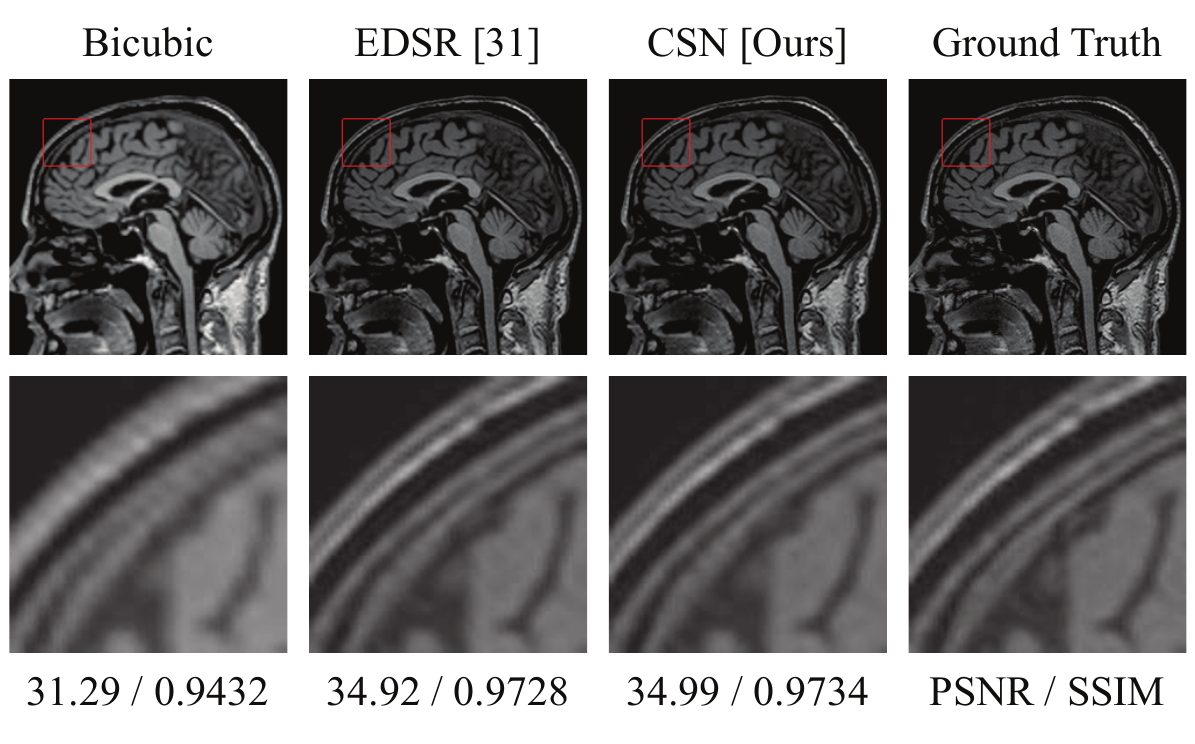}
  \end{minipage}}
  \subfigure[$k$-space truncation (TD)]{\label{fig11b}
  \begin{minipage}[t]{0.48\textwidth}
    \centering
    \includegraphics[scale = 0.68]{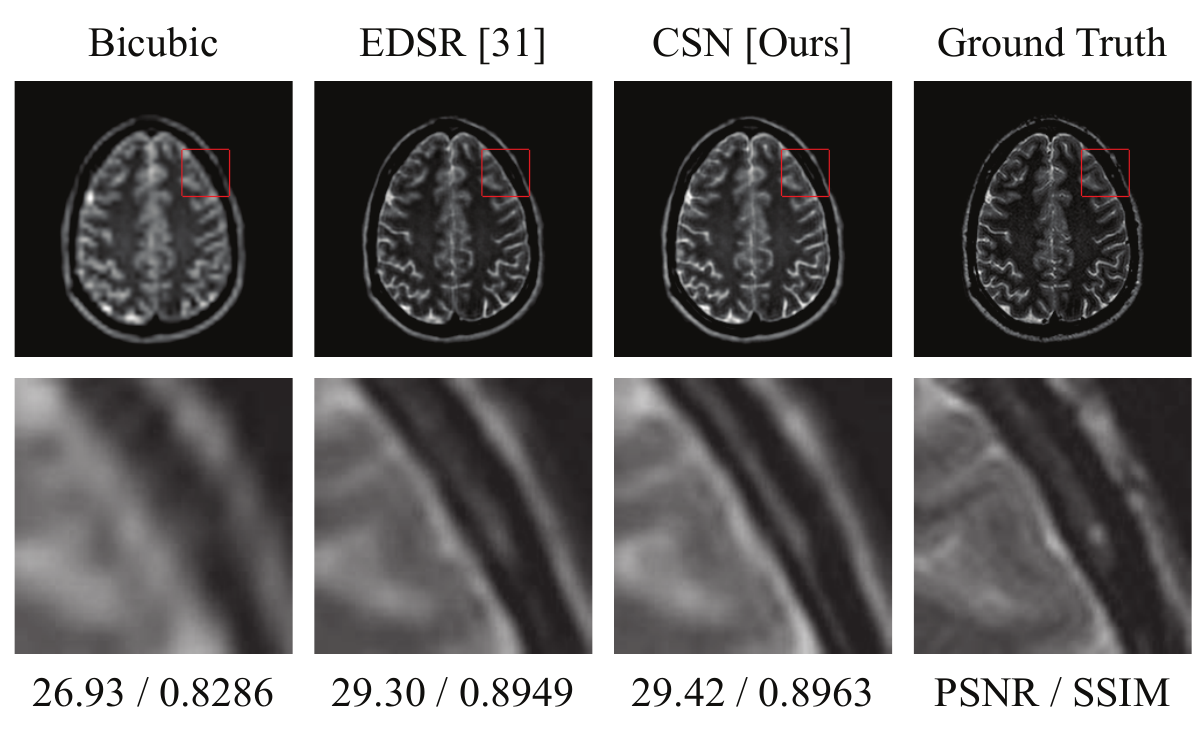}
  \end{minipage}}
  \vspace{-2mm}
  \caption{The visual comparison of EDSR \cite{Lim2017Enhanced} and the proposed CSN model in case of C96. (a) a T1 image with scaling factor SR$\times$2. (b) a T2 image with scaling factor SR$\times$4. In this case, each testing example is a 3D volume. For both (a) and (b), a randomly selected slice is used for display purposes.}
  \label{fig11}
\end{figure*}

\subsubsection{Bicubic Degradation (BD)}
Bicubic downsampling is a widely used simulation of LR image generation in image SR settings which simply shrinks HR images to a smaller size with the bicubic kernel. We examine this image degradation in this section first. Columns 4 to 6 of Table \ref{tab:C1} show that the quantitative results of the compared methods over the testing datasets under this image degradation. Overall, all the deep learning based methods (SRCNN \cite{Dong2016Image}, VDSR \cite{Kim2016Accurate}, RDN \cite{Zhang2018Residual}, CMSCN \cite{Hu2018Single}, FSCWRN \cite{Shi2019MR}, EDSR \cite{Lim2017Enhanced} and our CSN) present great advantages over the traditional methods (Bicubic and NLM \cite{Manjon2010Nonlocal}). However, the proposed CSN model gives the best SR performance in both C1 and C96 cases although it has fewer parameters and shallower model structures than RDN \cite{Zhang2018Residual} and EDSR \cite{Lim2017Enhanced}.

Fig.\ref{fig9} displays the visual comparison of these methods under the bicubic degradation. The top row shows the result on a PD image with scaling factor SR$\times$3. It can be observed that our CSN model has successfully restored the black area and it has the most similar shape to the ground truth. However, several other methods, such as Bicubic, NLM \cite{Manjon2010Nonlocal}, SRCNN \cite{Dong2016Image} and even VDSR \cite{Kim2016Accurate} almost lost the black area. The middle row is the result on a T1 image with SR$\times$4. There is a gray ridge at the position marked by the {\textcolor[rgb]{1,0,0}{red arrow}}, which can hardly be recognized in the results of other methods, but our model gives a more credible indication of the ridge. Similar results can also be observed from the bottom row, which shows the comparison on a T2 image with SR$\times$4. There is a dark ditch at the position marked by the {\textcolor[rgb]{1,0,0}{red arrow}}, and only our CSN model has succeeded in restoring this information.

\subsubsection{Truncation Degradation (TD)}
$k$-space truncation of HR images is a process that simulates the real image acquisition process where a LR image is scanned by reducing acquisition lines in both phase and slice encoding directions. The missing information is therefore in $k$-space and the degradation pattern of LR images is different from simply shrinking the size of HR images in the image domain by, e.g., bicubic interpolation \cite{Chen2018Brain}. Table \ref{tab:C1} also shows the quantitative comparison between different methods under the truncation degradation. Again, the proposed CSN model presents the best SR performance in both C1 and C96 cases. It is worth noting that the performance of Bicubic, NLM \cite{Manjon2010Nonlocal}, SRCNN \cite{Dong2016Image} and VDSR \cite{Kim2016Accurate} is slightly worse than that of these methods in case of BD, e.g., SR$\times$2 on T2 images. On the contrary, the performance of other methods in case of TD is better than that in case of BD. This is probably because TD degrades image quality more seriously than BD and models such as RDN \cite{Zhang2018Residual}, CMSCN \cite{Hu2018Single},  FSCWRN \cite{Shi2019MR} and our CSN have better representational capacity than models such as SRCNN \cite{Dong2016Image} and VDSR \cite{Kim2016Accurate}.

Fig.\ref{fig10} presents the visual effects of the compared methods in case of TD and the proposed CSN model presents obvious advantages over other models. For instance, the bottom row is the results on a T2 image with SR$\times$4. Our CSN model is able to reconstruct the dark contour at the position indicated by the {\textcolor[rgb]{1,0,0}{red arrow}}, which cannot be clearly observed in the results of other models. The top and middle rows show the results on a PD and a T1 image, also highlighting the advantages of the proposed CSN model. Fig.\ref{fig11} shows the visual comparison between the EDSR \cite{Lim2017Enhanced} and our CSN (C96) model in case of C96. Both BD and TD are presented. In this case, our model has less obvious advantages over EDSR \cite{Lim2017Enhanced}, but still performs better on the whole.

\section{Discussion and Future Work}\label{sec:discussion}
\subsection{Multiple Branches}
Like the original MAR mapping in \cite{Zhao2017Deep}, the stage mapping in our CSB can also be easily extended to multiple branches ($\geqslant$3). The difference is that we branch the network by channel splitting, instead of feature reuse. In extreme cases, it can be extended to $c$ branches with each branch occupying one channel of the input feature. This means that we explicitly differentiate the hierarchical features rather than having the network learn to distinguish between different features. Therefore, when the training samples are degraded and the model is complex, it helps to ease model training.

\subsection{Depth and Width}
Branching the network by reusing the entire feature tensor makes the model much wider, like \cite{Zhao2017Deep,Hu2018Single}. This significantly increases model parameters when the network depth is the same. Since EDSR \cite{Lim2017Enhanced} is a typical network with very wide structure and causes training failure, while RDN \cite{Zhang2018Residual} is a deeper but less wide network and can be successfully trained. Therefore, we speculate that the width of the model may also be one of the reasons for training failure in case of training sample degradation. Our work can be regarded as a manner to going deeper with nearly unchanged model width and parameters.

\subsection{Branch Structure}
Currently, we only utilize the structures similar to ResNet \cite{He2015Identity,He2016Deep} and DenseNet \cite{Huang2016Densely} (or RDN \cite{Zhang2018Residual}) for different branches. The experimental results show that mixing different branch structures is helpful to improve the performance of the model, but it is not conspicuous. This is probably because the structural difference between the two branches is relatively small. We conjecture that as the structural difference of the branches increases, so does the performance difference. The further investigation will be a part of our future work.

\subsection{3D Extension}
The present work only aims at the task of 2D MR image super-resolution, and the further extension could be in 3D case. However, since many types of medical images are in 3D format, it is intuitively possible to further enhance SR performance if the 3D structural information can be reasonably utilized \cite{Hu2016Single,Pham2017Brain,Chen2018Brain,Chen2018Efficient}. A prominent problem in the 3D settings is that the number of parameters will increase dramatically as the network depth increases, leading model training more difficult. Our model can deepen the network without significantly increasing the model width and parameters, which also helps to extend 3D models.

\subsection{Information Sharing}
In this paper, we only deal with the SR task for a single type of 2D MR images and a single scaling factor. However, there is evidence that combining the information from different image types and scaling factors is helpful to improve the performance of deep models \cite{Kim2016Accurate,Lim2017Enhanced}. The medical images SR framework combined multi-type and multi-scale information is also expected to further improve the SR performance of deep models.

\section{Conclusion}
\label{sec:conclusion}
A major problem with using deep models to super-resolve MR images is the lack of \textit{high-quality} and \textit{effective} training samples, which probably leads to performance degradation or even training failure of deep models. In this work, we have presented a novel deep channel splitting network (CSN) for the task of 2D MR image super-resolution, which is primarily made up of a series of cascaded channel splitting blocks (CSBs). The hierarchical features are split into two branches with different information propagations (residual branch and dense branch), which helps the model to discriminate different features explicitly. To integrate branch information, the MAR \cite{Zhao2017Deep} mapping is also applied to merge the hierarchical features on different branches.

Channel splitting helps to increase the depth of the network and the diversity of processing the hierarchical features. We conjecture that the performance improvement of the proposed model benefits from both two parts and additional performance can be further gained by exploring other branch structures and information fusion strategies. As it improves the dilemma between improving model performance and easing model training to some extent, it also has the potential to deal with other types of medical images, such as CT, ultrasound and PET etc.





\ifCLASSOPTIONcaptionsoff
  \newpage
\fi

\end{document}